\documentclass[runningheads]{llncs}

\usepackage{eccv}

\usepackage{eccvabbrv}

\usepackage{graphicx}
\usepackage{booktabs}
\usepackage{amsmath}
\usepackage{amssymb}
\usepackage{array}
\usepackage{lipsum}
\usepackage{cases}
\usepackage{bm}
\usepackage{makecell}
\usepackage{multirow}
\usepackage{color}
\usepackage{colortbl}
\definecolor{mygray}{gray}{.9}

\usepackage{appendix}

\usepackage[ruled,vlined]{algorithm2e}
\definecolor{commentcolor}{RGB}{110,154,155}   
\newcommand{\PyComment}[1]{\ttfamily\textcolor{commentcolor}{\# #1}}  
\newcommand{\PyCode}[1]{\ttfamily\textcolor{black}{#1}} 

\usepackage[accsupp]{axessibility}  

\usepackage[pagebackref,breaklinks,colorlinks,citecolor=eccvblue]{hyperref}

\usepackage{orcidlink}

\begin{document}

\title{EA-VTR: Event-Aware Video-Text Retrieval} 

\titlerunning{EA-VTR}

\author{Zongyang Ma\inst{1,2,3*} \and
Ziqi Zhang\inst{1*}\textsuperscript{$\dagger$} \and
Yuxin Chen\inst{1,2,3*} \and 
Zhongang Qi\inst{2} \and
Chunfeng Yuan\inst{1} \and
Bing Li\inst{1} \and
Yingmin Luo\inst{2} \and
Xu Li\inst{2} \and
Xiaojuan Qi\inst{5} \and
Ying Shan\inst{2} \and
Weiming Hu\inst{1,3,4}
}

\authorrunning{Z. Ma et al.}

\institute{MAIS, Institute of Automation, Chinese Academy of Sciences \email{\{mazongyang2020@,chenyuxin2019\}ia.ac.cn \{ziqi.zhang,cfyuan,bli,wmhu\}@nlpr.ia.ac.cn}\and
ARC Lab, Tencent PCG \email{\{zhongangqi,yingminluo,nelsonxli,yingsshan\}@tencent.com} \and
School of Artificial Intelligence, University of Chinese Academy of Sciences \and
School of Information Science and Technology, ShanghaiTech University \and
The University of Hong Kong \email{\{xjqi@eee.hku.hk\}}}

\maketitle

\begin{abstract}
Understanding the content of events occurring in the video and their inherent temporal logic is crucial for video-text retrieval. 
However, web-crawled pre-training datasets often lack sufficient event information, and the widely adopted video-level cross-modal contrastive learning also struggles to capture detailed and complex video-text event alignment. 
To address these challenges, we make improvements from both data and model perspectives.
In terms of pre-training data, we focus on supplementing the missing specific event content and event temporal transitions with the proposed event augmentation strategies. 
Based on the event-augmented data, we construct a novel Event-Aware Video-Text Retrieval model, \ie, EA-VTR, which achieves powerful video-text retrieval ability through superior video event awareness. 
EA-VTR can efficiently encode frame-level and video-level visual representations simultaneously, enabling detailed event content and complex event temporal cross-modal alignment, ultimately enhancing the comprehensive understanding of video events.
Our method not only significantly outperforms existing approaches on multiple datasets for Text-to-Video Retrieval and Video Action Recognition tasks, but also demonstrates superior event content perceive ability on Multi-event Video-Text Retrieval and Video Moment Retrieval tasks, as well as outstanding event temporal logic understanding ability on Test of Time task.
\end{abstract}

\begingroup
\renewcommand\thefootnote{}
\footnotetext{* Equal contribution. $\dagger$ Corresponding author.}
\endgroup

\section{Introduction}
\label{sec:intro}
\begin{figure}
    \centering
    \includegraphics[width=\linewidth]{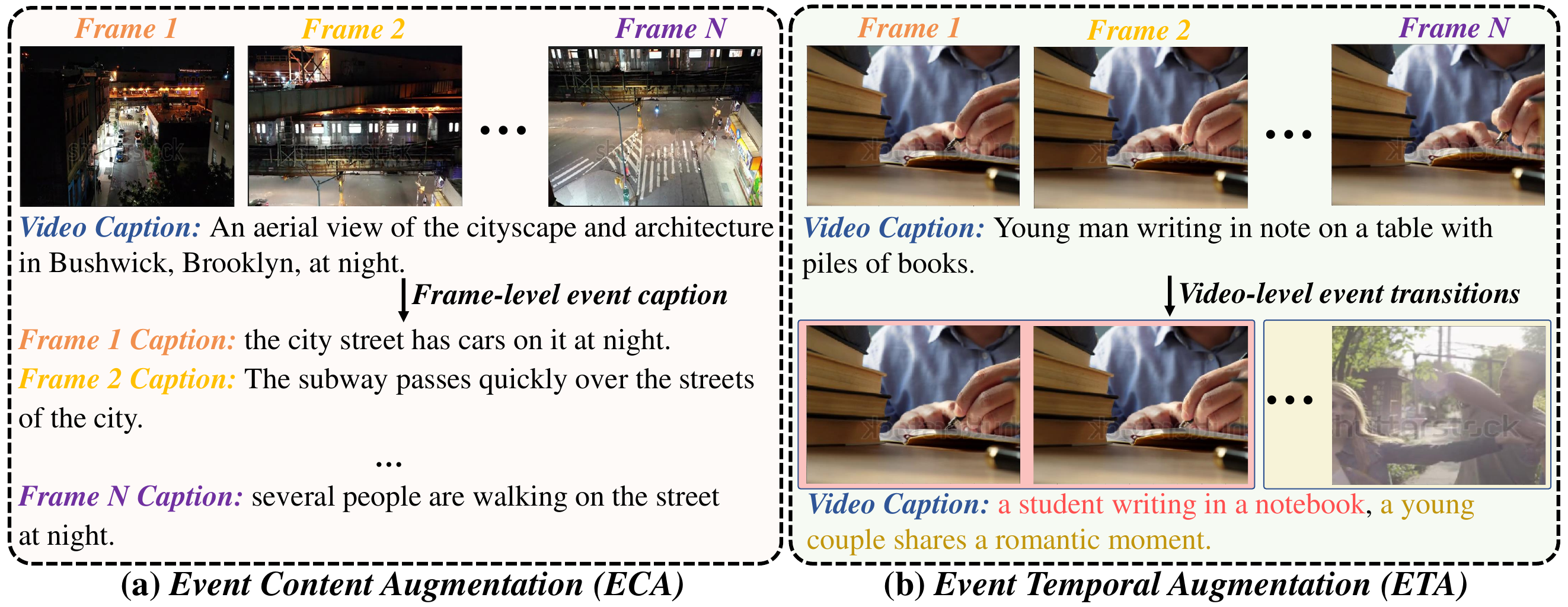}
    \caption{
    Examples of missing (a) event content and (b) temporal transitions and corresponding augmentation results.
    First, the web-crawled video caption in (a) does not contain specific event content.
    Second, in a video-text pair like (b), the video either lacks event temporal transitions or the caption does not reflect these transitions. 
    Therefore, we propose ECA and ETA to supplement the missing information in both aspects.
    }
    \vspace{-0.4cm}
    \label{fig:intro}
\end{figure}

The emergence of web-crawled pre-training datasets \cite{sharma2018conceptual, miech2019howto100m, bain2021frozen} has spurred rapid advancements in the field of video-text retrieval. 
Among the two types of retrieval models, \ie, joint-encoder \cite{zhu2020actbert, li2020hero, fu2021violet, li2022align} and dual-encoder \cite{bain2021frozen, ge2022bridging, ge2022miles, wang2022object, shi2023learning} models, the latter can achieve real-time retrieval, thus attracting more attention and providing wider practical applications.
Enhancing the ability to associate fine-grained video-text information is an effective way to improve the performance of dual-encoder models. 
While existing works have explored various details such as objects \cite{ge2022bridging, wang2022object}, actions \cite{ge2022bridging} and regions \cite{yan2023video}, the study of video events has been overlooked.
A video event refers to the visual activity or scene that occurs within a specific time interval in the video, which is a fundamental component of video and a concept widely used in video understanding works \cite{Shao_2018_ECCV, zhang2023multi}.
A video consists of one or more events, as illustrated in Figure \ref{fig:intro} (a), where the events \textit{``...  street has cars ...''}, \textit{``... subway passes...''}, and \textit{``... people are walking ...''} transpire respectively in the Frames 1, 2, and \textit{N} of the video.

Both the current pre-training datasets and elaborated models restrict the ultimate event understanding ability.
In terms of data, mainstream pre-training datasets, such as WebVid \cite{bain2021frozen}, lack captions about specific event content, \eg, the video caption ``An aerial view ...'' in Figure \ref{fig:intro} (a) does not cover the event \textit{``... subway passes ...''} in Frame 2. 
SViTT \cite{li2023svitt} also notes that many WebVid captions are associated with static backgrounds rather than foreground events.
Furthermore, a considerable portion of video-text pairs do not exhibit obvious event temporal transitions, as the video at the top of Figure \ref{fig:intro} (b) only contains a single event. 
To support this point, we have counted that 28\% of sampled 10,000 WebVid videos have only one event, by calculating whether any video frame has a similarity lower than the pre-defined threshold with the first frame (details in supplementary material).
Similarly, SViTT \cite{li2023svitt} claims that some WebVid videos consist of only simple motions without obviously dynamic changes.
In terms of the models, the commonly used video-level contrastive learning directly aligns global video and text representations, thus falls short in capturing detailed event content and complex event temporal alignment simultaneously, as evidenced by the comparisons in Table \ref{tab:other} and the relevant analysis.

To achieve a video-text retriever with robust event comprehension, we focus on enriching the event information in the pre-training data and improving the model's ability to perform event cross-modal alignment.
To obtain the content of multiple events, we first divide the video into several clips uniformly (consistent with the video encoder in video-text retrieval models), and then extract one frame from each clip and get the frame caption through an image captioner to capture the event content of the clip. 
This approach is based on a validated assumption through our manual verification, which posits that short videos inherently exhibit limited dynamic changes, and thus the divided clips present even less obvious dynamic changes, allowing a single frame to represent the clip adequately.
We refer to this process as Event Content Augmentation (ECA) and the demonstration is shown in Figure \ref{fig:intro} (a) for the captions \textit{``the city ...''},  \textit{``the subway ...''} and  \textit{``several people ...''} corresponding to Frames 1, 2, and \textit{N}. 
To construct video-text pairs with explicit event temporal transitions to train the model's event temporal understanding ability, we randomly intercept consecutive frames from two videos to synthesize into a new video, and the intercepted frame captions are also concatenated as the video caption, resulting in a synthesized video-text pair with clear event transitions.
We refer to the above process as Event Temporal Augmentation (ETA) and give an example in Figure \ref{fig:intro} (b).

Based on the event-augmented pre-training data obtained by applying ECA and ETA, we further introduce a novel Event-Aware Video-Text Retrieval model, \ie, EA-VTR, which emphasizes Event Content Learning (ECL) and Event Temporal Learning (ETL) in addition to the common video-level contrastive learning.
EA-VTR starts with adding several Frame [CLS] tokens to the video encoder, which interact with the visual patches of the corresponding frames to aggregate intra-frame information, allowing EA-VTR to compute both video-level and frame-level visual features with a negligible increase in computational overhead.
ECL then aligns frame features with the text features of the frame captions obtained through ECA to enhance the understanding of event content, and ETL aligns video-text features of the temporally rich synthesized video-text pairs obtained through ETA to improve event temporal logic understanding.
During the test, only video-level representation is taken to calculate similarity with the text query, thus ensuring efficient retrieval.

We first evaluate our method on text-to-video retrieval task, and the results demonstrate that our method not only significantly outperforms the best dual-encoder methods on various datasets, such as achieving a 4.7\% and 5.3\% improvement in R@10 on MSRVTT and DiDeMo under zero-shot settings, but also exhibit comparable performance to the SOTA joint-encoder methods with over 30x speed advantage. 
Furthermore, our method generalizes well on video action recognition tasks, with a 7.9\% increase in mean accuracy on UCF101 compared to the best method. 
Moreover, we thoroughly evaluate the event understanding ability of our method, including Multi-event Video-Text Retrieval  \cite{zhang2023multi} and Video Moment Retrieval tasks that involve perceiving all events and specific events within a video, as well as Test of Time task \cite{bagad2023test} that focus on understanding the temporal logic between multiple events. The results demonstrate that our method outperforms previous approaches in all three tasks, encompassing both content and temporal logic aspects of event understanding.
The contributions of this work are listed as follows:
\begin{itemize}
    \item
    We propose Event Content Augmentation and Event Temporal Augmentation to complement the event content and temporal transitions lacking in the pre-training data.
    \item
    We introduce a novel Event-Aware Video-Text Retrieval model, which achieves superior cross-modal retrieval and event understanding abilities by performing Event Content Learning and Event Temporal Learning. 
    \item
    The results of Text-to-Video Retrieval and Video Action Recognition tasks show our effectiveness in video-text retrieval, while the results of Multi-event Video-Text Retrieval, Video Moment Retrieval, and Test of Time tasks further validate our superiority in video event understanding. 
\end{itemize}

\section{Related Work}
 
\noindent \textbf{Pre-training for Video-Text Retrieval.}
Previous works for video-text retrieval can be divided into joint-encoder \cite{zhu2020actbert, li2020hero, fu2021violet, li2022align, li2023lavender, wang2023all, wu2022rap, huang2023clover, Yang_2023_ICCV, ye2023hitea} and dual-encoder \cite{liu2019use, patrick2020support, miech2020end, gabeur2020multi, rouditchenko2020avlnet, ging2020coot, bain2021frozen, xu2021videoclip, yang2021taco, ge2022bridging, ge2022miles, bai2022lat, wang2022object, yan2023video} models. 
The consensus within the cross-modal retrieval community \cite{Lu_2022_CVPR, ge2022bridging, Chen_2023_CVPR, Chen_2024_CVPR} is that joint-encoder models perform cross-modal interactions at the expense of speed to achieve higher accuracy, whereas dual-encoder models adopt two individual encoders to extract video \cite{bertasius2021space, bain2021frozen, liu2022video} and text \cite{devlin2018bert, sanh2019distilbert} representations separately and then calculate their cross-modal cosine similarities, enabling efficient real-time retrieval and gaining wide practical application.
To improve the generalization of efficient dual-encoder models on downstream tasks, some works \cite{ge2022bridging,wang2022object,yan2023video} have explored improving their ability to associate fine-grained video-text information. 
MCQ \cite{ge2022bridging} constructs multiple-choice questions about objects and actions in the video and forces the model to select the correct option. 
OA-Trans \cite{wang2022object} aligns object regions, obtained through the object detector \cite{ren2015faster}, with their corresponding labels.
RegionLearner \cite{yan2023video} clusters similar visual patches to approximate objects to align with text representations. 
However, current works overlook the study of video events, which are fundamental components of videos.
Understanding events not only directly benefits tasks that require event perception \cite{zhang2023multi} and logical judgment \cite{bagad2023test}, but also holds the potential to improve cross-modal retrieval performance by enhancing the grasp of video details.
To unleash the potential of dual-encoder models, we are committed to enhancing the event understanding ability of the retriever from both data and model perspectives in this work.

\noindent \textbf{Data Augmentation for Video-Text Pre-Training.} 
There are some works \cite{xue2022clip, wu2023cap4video, zhao2023learning,xu2021boundary} devoted to augmenting the video or text information in the training data to obtain improved video-text pre-training models.
In terms of text, CLIP-ViP claims the language domain gap between text in pre-training and downstream datasets will reduce the generalization ability, and thus use an image-text pre-training model \cite{wang2022ofa} to generate single-frame captions for HowTo100M \cite{miech2019howto100m}.
Cap4Video utilizes ZeroCap \cite{tewel2022zerocap}, which combines CLIP \cite{radford2021learning} and GPT2 \cite{radford2019language}, to create a caption for each video, simulating the real-world scenario where videos are accompanied by related textual information (such as titles and tags) for retrieval.
LAVILA \cite{zhao2023learning} generates additional narrations for long videos from the Ego4D dataset \cite{ego4degocentric, grauman2022ego4d} through a fine-tuned large language model \cite{radford2019language}. 
On the video front, BSP \cite{xu2021boundary} employs a strategy of fusing two videos to synthesize a new video for temporal localization pre-training.
In contrast to the unimodal data augmentation in the aforementioned works, we perform data augmentation on both video and text modalities to comprehensively supplement event content and temporal transition information in pre-training data.

\section{Method}

In this section, we first introduce the proposed event augmentation strategy in section \ref{sec:dataaug}.
Then we present the novel Event-Aware Video-Text Retrieval model in section \ref{sec:ea-vtr}.
Finally, we give the model training and inference in section \ref{sec:train_infer}.  

\begin{figure*}
    \centering
    \includegraphics[width=\textwidth]{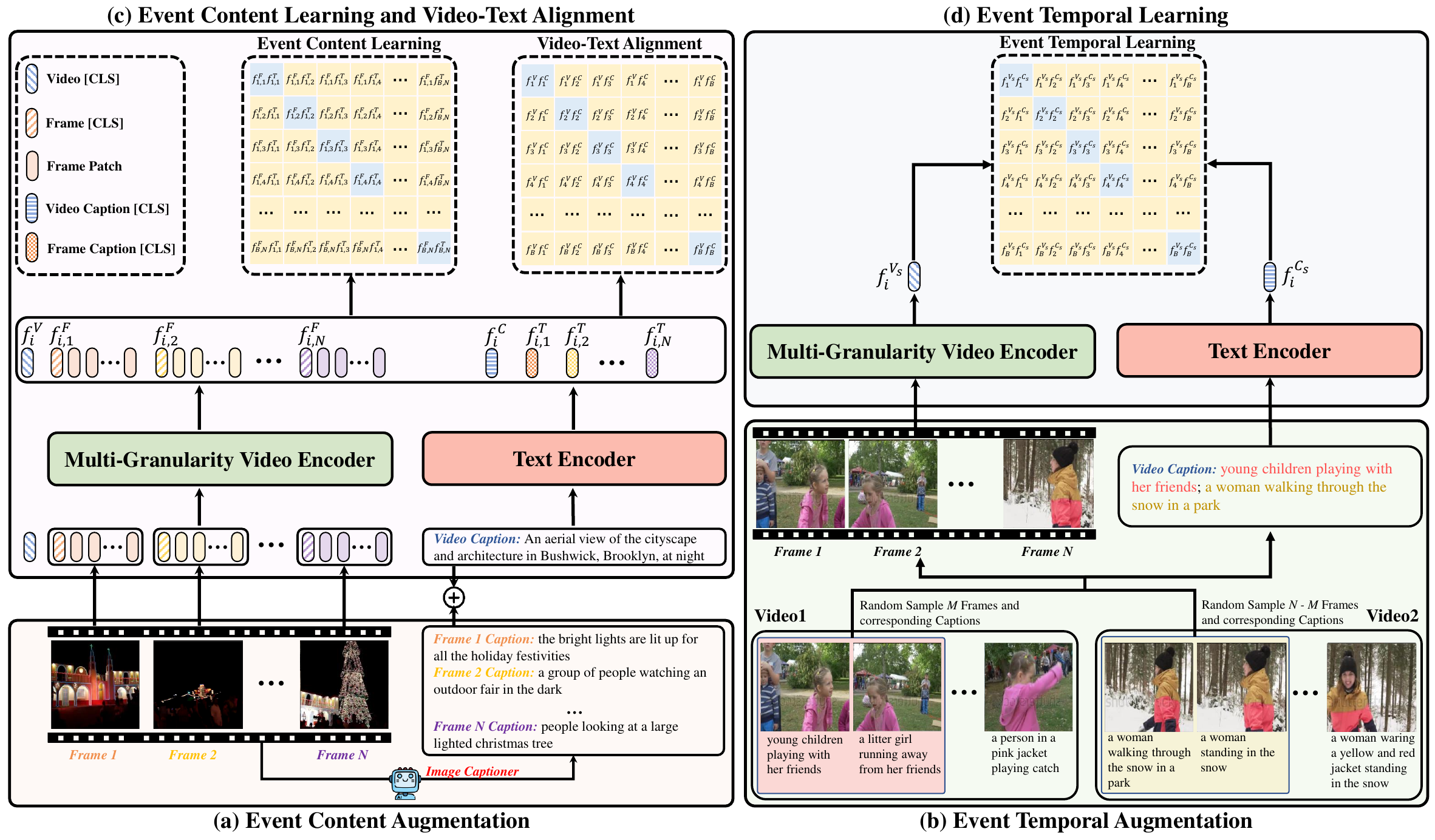}
    \vspace{-0.3cm}
    \caption{
    Overview of the proposed Event Content Augmentation (a) and Event Temporal Augmentation (b) to augment the event information in the pre-training dataset, and EA-VTR model using Event Content Learning (c) and Event Temporal Learning (d) to learn from the augmented data. 
    }
    \label{fig:model}
    \vspace{-0.3cm}
\end{figure*}

\subsection{Event Augmentation for Pre-Training}
\label{sec:dataaug}
Event augmentation aims to provide data with both rich event content and temporal transitions for model pre-training, which includes Event Content Augmentation and Event Temporal Augmentation. 

\subsubsection{3.1.1 Event Content Augmentation}
\label{sec:eca}
As shown in Figure \ref{fig:model} (a), Event Content Augmentation (ECA) captures the content of events occurring in different clips (or different segments, different durations) of the video $V_i$ with web-crawled caption $C_i$.
We first uniformly divide the video $V_i$ into $N$ non-overlapping clips, and then sample a single frame from each clip to form the frame set $F_i = \{F_{i,j}\}_{j=1}^N$, which is the same input strategy as the video encoder in the retrieval models \cite{bain2021frozen, ge2022bridging, ge2022miles}. 
The sampled frames $F_i$ are then fed into an off-the-shelf image captioner to obtain the corresponding frame captions $T_i = \{T_{i,j}\}_{j=1}^N$, which are used to capture the event content within each clip.
The feasibility of using a single frame to represent a clip of short video has been validated through our manual verification.
To generate more diverse frame captions for training a model with strong generalization ability, we employ a non-deterministic Top-p (Nucleus) sampling \cite{holtzman2019curious} generation strategy, ultimately equipping the video $V_i$ with $N$ frame-level image-text pairs.

\subsubsection{3.1.2 Event Temporal Augmentation}
\label{sec:eta}
The process of Event Temporal Augmentation (ETA) is shown in Figure \ref{fig:model} (b).
In a batch, two randomly selected videos, named Video1 and Video2, have $M_1$ and $M_2 \ (M_1, M_2 \leq N)$ frames randomly extracted from multiple consecutive video clips, denoted as $F_1 = \{F_{1,j}\}_{j=1}^{M_1}$ and $F_2 = \{F_{2,j}\}_{j=1}^{M_2}$. 
The corresponding frame captions, $T_1 = \{T_{1,j}\}_{j=1}^{M_1}$ and $T_2 = \{T_{2,j}\}_{j=1}^{M_2}$, are also selected.
Then we have tried two video synthesis schemes, Video Weighted Summation and Video Concatenation.

\noindent \textbf{Video Weighted Summation.}
This scheme requires the frame numbers to meet $M_1 = M_2 = N$, and a newly synthesized video $V^{s}_i$ is obtained by taking the weighted summation of the original pixels of $F_1$ and $F_2$:
\vspace{-0.1cm}
\begin{equation}
\vspace{-0.1cm}
    V_i^s = {\rm Concat}(\{w_j \cdot F_{1,j} + (1 - w_j) \cdot F_{2,j}\}_{j=1}^N),
\end{equation}
where weights $w_j$ should be monotonically decreasing, ensuring that the synthesized video $V^{s}_i$ primarily focuses on the events of Video1 in the first half and the events of Video2 in the second half.

\noindent \textbf{Video Concatenation.}
In this scheme, the frame numbers $M_1$ and $M_2$ are supposed to satisfy $M_1 + M_2 = N$, and a newly synthesized video $V^{s}_i$ is obtained by concatenating all frames of $F_1$ and $F_2$:
\vspace{-0.1cm}
\begin{equation}
\vspace{-0.1cm}
    V_i^s = {\rm Concat}(F_1, F_2).
\end{equation}

\noindent \textbf{Caption for the synthesized video.} 
The accompanying caption $C_i^s$ of synthesized video $V^{s}_i$ is construct by concatenating two frame captions sampled from $T_1$ and $T_2$ respectively.

Since the synthesized video contains evident event transition changes, the video events exhibit distinct separability in the temporal dimension, which requires the model to understand the sequential logic between events to effectively comprehend the synthesized video.

\subsection{Event-Aware Video-Text Retrieval}
\label{sec:ea-vtr}
With the foundation of the aforementioned augmented pre-training data, the novel Event-Aware Video-Text Retrieval model can be unfolded according to the following outline.

\subsubsection{3.2.1 Multi-Granularity Video Encoder}
\label{sec:multi-grani}
Since the pre-training data augmented by ECA and ETA contains both frame-level image-text pairs as well as web-crawled and synthesized video-level video-text pairs, this naturally requires the video encoder to output both video-level and frame-level visual representations to enable subsequent cross-modal alignment at multi-granularity. 
However, the video encoders of most current video-text retrieval models are TimeSformer \cite{bertasius2021space} or its variants \cite{bain2021frozen, ge2022miles}, which only support providing video-level visual representations. 

To address this issue, we make a simple but generic modification to TimeSformer or its variants \cite{bain2021frozen, ge2022miles}, enabling it to provide video and frame representations simultaneously.
As shown in Figure \ref{fig:model} (c), in contrast to the previous video encoder, we concatenate a new Frame [CLS] token embedding $e^F_{i,j}$ with random initialization before the frame patch embeddings $e^P_{i,j}$ of the video frame $F_{i,j}$.
After incorporating the spatio-temporal position information into the aforementioned embeddings, a video [CLS] embedding $e^V_i$ is finally concatenated at the beginning, and the resulting embeddings $e_i = {\rm Concat}(e^V_i; \{e^F_{i,j}; e^P_{i,j}\}_{j=1}^N)$ are then fed into the video encoder $E_V$ to be converted into representations $f_i$:
\vspace{-0.1cm}
\begin{equation}
\vspace{-0.1cm}
    f_i = E_V(e_i),
\end{equation}
where $f_i = {\rm Concat}(f^V_i; \{f^F_{i,j}; f^P_{i,j}\}_{j=1}^N)$, thus we can directly extract video representation $f^V_i$ and frame representations $\{f^F_{i,j}\}_{j=1}^N$.
$E_V$ adopts the similar architecture as \cite{bain2021frozen}. 
On the one hand, this enables the new Frame [CLS] tokens to primarily interact with the visual patches within the respective frames to form frame representations. 
On the other hand, the additional computational cost introduced by the $N$ new Frame [CLS] tokens (since $N$ is much smaller than the number of visual patches) is negligible.

\subsubsection{3.2.2 Event Content Learning}
\label{sec:ecl}
We first perform Event Content Learning (ECL) by aligning video frames with the corresponding frame captions to capture the detailed event content of the video.
As shown in Figure \ref{fig:model} (c), each frame caption in $\{T_{i,j}\}_{j=1}^N$ is individually fed into the text encoder $E_T$, yielding the corresponding text features $\{f^T_{i,j}\}_{j=1}^N$. 
Subsequently, we employ contrastive learning \cite{jozefowicz2016exploring, oord2018representation} to align the frame visual features with the frame caption features, which is defined as follows:
\vspace{-0.1cm}
\begin{equation}
\vspace{-0.1cm}
    \begin{aligned}
       \mathcal{L}^{F2T}_{i,j} &= - log \frac{\mathrm{exp}(f^F_{i,j}, f^T_{i,j}, \tau_c)}{\sum_{i^{'}=1}^B \sum_{j^{'}=1}^N \mathrm{exp}(f^F_{i,j}, f^T_{i^{'},j^{'}}, \tau_c)}, \\
       \mathcal{L}^{T2F}_{i,j} &= - log \frac{\mathrm{exp}(f^T_{i,j}, f^F_{i,j}, \tau_c)}{\sum_{i^{'}=1}^B \sum_{j^{'}=1}^N \mathrm{exp}(f^T_{i,j}, f^F_{i^{'},j^{'}}, \tau_c)}, \\
    \end{aligned}
\end{equation}
where $\mathrm{exp}(x,y,\tau)=e^{x^{\top}y/\tau}$, $\tau_c$ is a temperature hyper-parameter, and $B$ is the batch size. 
The total loss for ECL is defined as:
\vspace{-0.1cm}
\begin{equation}
\vspace{-0.1cm}
    \mathcal{L}_{ECL} = \frac{1}{B \cdot N} \sum_{i=1}^B \sum_{j=1}^N  (\mathcal{L}^{F2T}_{i,j} + \mathcal{L}^{T2F}_{i,j}) / 2.
\end{equation}

\subsubsection{3.2.3 Event Temporal Learning}
\label{sec:etl}
We perform Event Temporal Learning (ETL) by aligning the temporally rich synthesized video and the paired caption. 
From the process depicted in Figure \ref{fig:model} (d), we can see that the synthesized video $V^S_i$ and its corresponding caption $C^S_i$ are separately passed through the video encoder $E_V$ and text encoder $E_T$, resulting in their respective features $f^{V^s}_i$ and $f^{C^s}_i$.
We then still use contrastive learning to enable cross-modal alignment, which is defined as follows:
\vspace{-0.1cm}
\begin{equation}
\vspace{-0.1cm}
    \begin{aligned}
       \mathcal{L}^{V^s2C^s}_{i} &= - log \frac{\mathrm{exp}(f^{V^s}_{i}, f^{C^s}_{i}, \tau_t)}{\sum_{i^{'}=1}^B \mathrm{exp}(f^{V^s}_{i}, f^{C^s}_{i^{'}}, \tau_t)}, \\
       \mathcal{L}^{C^s2V^s}_{i} &= - log \frac{\mathrm{exp}(f^{C^s}_{i}, f^{V^s}_{i}, \tau_t)}{\sum_{i^{'}=1}^B \mathrm{exp}(f^{C^s}_{i}, f^{V^s}_{i^{'}}, \tau_t)}, \\
    \end{aligned}
\end{equation}
the total loss for ETL is defined as:
\vspace{-0.1cm}
\begin{equation}
\vspace{-0.1cm}
    \mathcal{L}_{ETL} = \frac{1}{B} \sum_{i=1}^B  (\mathcal{L}^{V^s2C^s}_{i} + \mathcal{L}^{C^s2V^s}_{i}) / 2.
\end{equation}

\subsection{Training and Inference}
\label{sec:train_infer}
\textbf{Training.} In addition to ECL and ETL, video-level contrastive learning is also employed to align the original video-text pair in training data, which we refer to as Video-Text Alignment (VTA), as shown in Figure \ref{fig:model} (c) and consistent with previous work.
The model is optimized through the joint constraints of ECL, ETL, and VTA losses.
However, we find that directly mixing original and synthesized video-text pairs within a batch can interfere with model training, as the high distinguishability (due to the different information density) between the two results in minimal contrastive gains between them.
To address this, we adopt an Alternating Iteration Training scheme to separately utilize the two sets of data for training, with details in Algorithm \ref{algo:ait}.

\begin{algorithm}[h]
\SetAlgoLined
\scriptsize
    \PyComment{E\_V, E\_T: Video Encoder and Text Encoder} \\
    \PyComment{V, C, T: Videos, Captions and Frame Captions} \\
    \PyComment{Vs, Cs: Synthesized Videos and Captions} \\
    \PyComment{l\_vta, l\_ecl, l\_etl: VTA, ECL and ETL loss} \\
    \PyCode{for V, C, T, Vs, Cs in loader:} \\
    \Indp   
        \PyComment{Training on Original Video-Text Pairs} \\
        \PyCode{f\_V, f\_F = E\_V(V)} \\
        \PyCode{f\_C, f\_T = E\_T(C), E\_T(T)} \\
        \PyCode{loss = l\_vta(f\_V, f\_C) + l\_ecl(f\_F, f\_T)} \\
        \PyCode{loss.backward()} \\
        \PyComment{Training on synthesized Video-Text Pairs} \\
        \PyCode{f\_Vs = E\_V(Vs)} \\
        \PyCode{f\_Cs = E\_T(Cs)} \\
        \PyCode{loss = l\_etl(f\_Vs, f\_Cs)} \\
        \PyCode{loss.backward()} \\
    \Indm 
\caption{Alternating Iteration Training}
\label{algo:ait}
\end{algorithm}

\noindent \textbf{Inference.} During inference, we only take the video-level representation from the video encoder to respond to text queries for efficient retrieval, just like a typical dual-encoder-based retriever \cite{bain2021frozen}.

\section{Experiments}

\subsection{Datasets and Evaluation Metrics}
\label{sec:data}
\noindent \textbf{Pre-Training Datasets.}
We follow recent works and use two datasets for pre-training: Conceptual Captions 3M (CC3M) \cite{sharma2018conceptual}, containing 3 million image-text pairs, and WebVid-2M \cite{bain2021frozen}, containing 2.5 million video-text pairs.

\noindent \textbf{Downstream Datasets and Evaluation Metrics.}
We conduct downstream \textbf{Text-to-Video Retrieval} evaluation on four widely used datasets: 
MSRVTT \cite{xu2016msr}, DiDeMo \cite{anne2017localizing}, LSMDC \cite{rohrbach2015dataset} and MSVD \cite{chen2011collecting}. 
Recall and Median Rank are used as retrieval evaluation metrics.
Since the query text of DiDeMo is a concatenation of captions corresponding to multiple video clips with temporal relationships in the time dimension, the evaluation results on DiDeMo to some extent reflect the ability of the model to capture event temporal logic.
Furthermore, we validate the generalization ability of our method on two \textbf{Video Action Recognition} datasets, UCF101 \cite{soomro2012ucf101} and HMDB51 \cite{kuehne2011hmdb}, as in previous works. 
Top-1 classification accuracy is used as action recognition evaluation metric.
To demonstrate the advantage of our method in understanding video event content, we conducted experiments on the newly proposed \textbf{Multi-event Video-Text Retrieval} \cite{zhang2023multi} and \textbf{Video Moment Retrieval} tasks. 
To showcase the superiority of our method in clarifying event temporal logic, we perform experiments on the newly proposed \textbf{Test of Time} \cite{bagad2023test} task. 
The above three event understanding tasks all involve making certain adjustments to the original ActivityNet \cite{caba2015activitynet} annotations to serve as task data.
Multi-event Video-Text Retrieval employs Recall@k-Average, Recall@k-One-Hit and Recall@k-All-Hit as evaluation metrics; video moment retrieval uses $\mathrm{R}@^{\theta}_n$ as the evaluation metric; and Test of Time adopts time-order consistency as the evaluation metric.
The details of these downstream datasets and evaluation metrics can be found in the supplemental material.

\subsection{Implementation Details}
EA-VTR adopts the BERTbase \cite{devlin2018bert} model as the text encoder and a TimeSformer-variant model \cite{bain2021frozen} initialized with ViT \cite{dosovitskiy2020image} weights pre-trained on ImageNet-21k \cite{deng2009imagenet} as the video encoder.
BLIP \cite{li2022blip} is utilized as the image captioner with a Top-p of 0.9 for sampling during generation.
Pre-training is divided into two stages following previous works.
In the first stage, we warm up EA-VTR by pre-training it for 10 epochs on CC3M and WebVid-2M (random sample 1 frame), with a batch size of 2048 and a peak learning rate of 1×$10^{-4}$. 
In the second stage, we perform 5 epoch pre-training constrained by ETA, ECL and ETL jointly on WebVid-2M with 4 random sampled frames from evenly divided clips, with a batch size of 1024 and a peak learning rate of 5×$10^{-5}$. 
All images and videos are resized to 224 × 224 as input for pre-training.
AdamW \cite{loshchilov2017decoupled} is utilized as the optimizer, and the learning rate follows a cosine annealing learning rate schedule.
Both pre-training stages are performed on 32 V100 GPUs.
More implementation details are presented in the supplementary material.

\subsection{Comparison with the State-of-the-Art}

\vspace{-0.5cm}
\begin{table}[]
\centering
\caption{Zero-shot text-to-video retrieval results on MSRVTT, DiDeMo, LSMDC, and MSVD. \textcolor{gray}{CLIP-ViP} uses $\times$100 our training data.}
\vspace{-0.2cm}
\resizebox{\textwidth}{!}{%
\begin{tabular}{ccccccccccccccccc}
\toprule
\multicolumn{1}{c|}{\multirow{2}{*}{Method}} & \multicolumn{4}{c|}{MSRVTT} & \multicolumn{4}{c|}{DiDeMo} & \multicolumn{4}{c|}{LSMDC} & \multicolumn{4}{c}{MSVD} \\
\multicolumn{1}{c|}{} & R@1$\uparrow$ &R@5 $\uparrow$ & R@10$\uparrow$ & \multicolumn{1}{c|}{MedR$\downarrow$} & R@1$\uparrow$ &R@5 $\uparrow$ & R@10$\uparrow$ & \multicolumn{1}{c|}{MedR$\downarrow$} & R@1$\uparrow$ & R@5$\uparrow$ & R@10$\uparrow$ & \multicolumn{1}{c|}{MedR$\downarrow$} & R@1$\uparrow$ & R@5$\uparrow$ & R@10$\uparrow$ & MedR$\downarrow$ \\ \midrule
\multicolumn{17}{c}{\textbf{Joint-Encoder}} \\ \midrule
\multicolumn{1}{c|}{TACO \cite{yang2021taco}} & 9.8 & 25.0 & 33.4 & \multicolumn{1}{c|}{29.0} & -- & -- & -- & \multicolumn{1}{c|}{--} & -- & -- & -- & \multicolumn{1}{c|}{--} & -- & -- & -- & -- \\
\multicolumn{1}{c|}{VIOLET \cite{fu2021violet}} & 25.9 & 49.5 & 59.7 & \multicolumn{1}{c|}{--} & 23.5 & 49.8 & 59.8 & \multicolumn{1}{c|}{--} & -- & -- & -- & \multicolumn{1}{c|}{--} & -- & -- & -- & -- \\
\multicolumn{1}{c|}{ALPRO \cite{li2022align}} & 24.1 & 44.7 & 55.4 & \multicolumn{1}{c|}{8.0} & 23.8 & 47.3 & 57.9 & \multicolumn{1}{c|}{6.0} & -- & -- & -- & \multicolumn{1}{c|}{--} & -- & -- & -- & -- \\
\multicolumn{1}{c|}{Rap \cite{wu2022rap}} & 28.9 & 47.5 & 56.8 & \multicolumn{1}{c|}{7.0} & 29.5 & 55.7 & 65.6 & \multicolumn{1}{c|}{4.0} & 12.8 & 26.6 & 33.4 & \multicolumn{1}{c|}{37.0} & \textbf{35.9} & \textbf{64.3} & \textbf{73.7} & \textbf{3.0} \\
\multicolumn{1}{c|}{Clover \cite{huang2023clover}} & 26.4 & 49.5 & \textbf{60.0} & \multicolumn{1}{c|}{6.0} & 29.5 & 55.2 & 66.3 & \multicolumn{1}{c|}{4.0} & \textbf{14.7} & 29.2 & \textbf{38.2} & \multicolumn{1}{c|}{\textbf{24.0}} & -- & -- & -- & -- \\
\multicolumn{1}{c|}{TW-BERT \cite{Yang_2023_ICCV}} & 26.4 & 50.1 & 59.6 & \multicolumn{1}{c|}{\textbf{5.0}} & 28.4 & 52.9 & 64.5 & \multicolumn{1}{c|}{\textbf{4.0}} & 14.2 & \textbf{30.4} & 36.0 & \multicolumn{1}{c|}{28.0} & -- & -- & -- & -- \\ 
\multicolumn{1}{c|}{Singularity \cite{lei2022revealing}} & \textbf{28.4} & \textbf{50.2} & 59.5 & \multicolumn{1}{c|}{--} & \textbf{36.9} & \textbf{52.9} & \textbf{64.5} & \multicolumn{1}{c|}{--} & -- & -- & -- & \multicolumn{1}{c|}{--} & -- & -- & -- & -- \\ \midrule
\multicolumn{17}{c}{\textbf{Dual-Encoder}} \\ \midrule
\multicolumn{1}{c|}{Frozen \cite{bain2021frozen}} & 18.7 & 39.5 & 51.6 & \multicolumn{1}{c|}{10.0} & 21.1 & 46.0 & 56.2 & \multicolumn{1}{c|}{7.0} & 9.3 & 22.0 & 30.1 & \multicolumn{1}{c|}{51.0} & 38.7 & 70.1 & 80.1 & 2.0 \\
\multicolumn{1}{c|}{LaT \cite{bai2022lat}} & 23.4 & 44.1 & 53.3 & \multicolumn{1}{c|}{8.0} & 22.6 & 45.9 & 58.9 & \multicolumn{1}{c|}{7.0} & - & - & -  &  \multicolumn{1}{c|}{-} & 36.9 & 68.6 & 81.0 & 2.0 \\
\multicolumn{1}{c|}{RegionL. \cite{yan2023video}} & 22.2 & 43.3 & 52.9 & \multicolumn{1}{c|}{8.0} & -- & -- & -- & \multicolumn{1}{c|}{--} & -- &  -- & -- & \multicolumn{1}{c|}{--} & -- & -- & -- & -- \\
\multicolumn{1}{c|}{OA-Trans \cite{wang2022object}} & 23.4 & 47.5 & 55.6 & \multicolumn{1}{c|}{8.0} & 23.5 & 50.4 & 59.8 & \multicolumn{1}{c|}{6.0} & - & - & - & \multicolumn{1}{c|}{-} & - & - & - & - \\
\multicolumn{1}{c|}{MCQ \cite{ge2022bridging}} & 26.0 & 46.4 & 56.4 & \multicolumn{1}{c|}{7.0} & 25.6 & 50.6 & 61.1 & \multicolumn{1}{c|}{5.0} & 12.2 & 25.9 & 32.2 & \multicolumn{1}{c|}{42.0} & 43.6 & 74.9 & 84.9 & 2.0 \\
\multicolumn{1}{c|}{Miles \cite{ge2022miles}} & 26.1 & 47.2 & 56.9 & \multicolumn{1}{c|}{7.0} & 27.2 & 50.3 & 63.6 & \multicolumn{1}{c|}{5.0} & 11.1 & 24.7 & 30.6 & \multicolumn{1}{c|}{50.7} & 44.4 & 76.2 & \textbf{87.0} & 2.0 \\ 
\multicolumn{1}{c|}{TCP \cite{Zhang_2023_ICCV}} & 26.8 & 48.3 & 57.6 & \multicolumn{1}{c|}{7.0} & -- & -- & -- & \multicolumn{1}{c|}{--} & -- & -- & -- & \multicolumn{1}{c|}{--} & -- & -- & -- & -- \\ 
\multicolumn{1}{c|}{\textcolor{gray}{CLIP-ViP \cite{xue2022clip}}} & \textcolor{gray}{31.7} & \textcolor{gray}{51.2} & \textcolor{gray}{63.2} & \multicolumn{1}{c|}{\textcolor{gray}{4.0}} & \textcolor{gray}{24.6} & \textcolor{gray}{50.7} & \textcolor{gray}{59.7} & \multicolumn{1}{c|}{\textcolor{gray}{5.0}} & \textcolor{gray}{12.5} & \textcolor{gray}{26.1} & \textcolor{gray}{33.3} & \multicolumn{1}{c|}{\textcolor{gray}{39.0}} & \textcolor{gray}{--} & \textcolor{gray}{--} & \textcolor{gray}{--} & \textcolor{gray}{--} \\ 
\rowcolor{mygray}
\multicolumn{1}{c|}{\textbf{EA-VTR}} & \textbf{28.0} & \textbf{53.1} & \textbf{62.3} & \multicolumn{1}{c|}{\textbf{5.0}} & \textbf{32.7} & \textbf{58.9} & \textbf{68.9} & \multicolumn{1}{c|}{\textbf{3.0}} & \textbf{15.7} & \textbf{29.6} & \textbf{36.0} & \multicolumn{1}{c|}{\textbf{30.0}} & \textbf{46.6} & \textbf{78.9} & 86.5 & \textbf{2.0} \\  \bottomrule
\end{tabular}%
}
\label{tab:zero-msrvtt-didemo-lsmdc-msvd}
\vspace{-0.5cm}
\end{table}

\noindent \textbf{Text-to-Video Retrieval.} We separately compare the results of joint-encoder and dual-encoder models as previous works \cite{Lu_2022_CVPR, Chen_2023_CVPR}. 
The zero-shot text-to-video retrieval results on four mainstream downstream datasets are presented in Table \ref{tab:zero-msrvtt-didemo-lsmdc-msvd}, we can draw the following conclusions:
(1) EA-VTR significantly surpasses previous dual-encoder methods by a large margin. 
Specifically, EA-VTR improves upon the previous best method TCP by 1.2\%, 4.8\%, and 4.7\% on R@1, R@5, and R@10 of MSRVTT. 
The advantage is even more pronounced on the temporally-rich DiDeMo, with R@1, R@5, and R@10 being 5.5\%, 8.6\%, and 5.3\% higher than the previous SOTA method Miles.
Even when compared to the dual-encoder method CLIP-ViP, which utilizes $\times$100 more training data than ours, our method still maintains competitiveness on MSRVTT and exhibits superior performance on DiDeMo and LSMDC. 
(2) Compared to the best joint-encoder methods on various datasets, EA-VTR consistently demonstrates comparable or even superior performance. Simultaneously, it has obvious speed advantages, such as being more than $\times$30 faster than Singularity \cite{lei2022revealing}.
This is noteworthy, as joint-encoder methods are typically considered to have superior performance but at the cost of lower efficiency.
The text-to-video retrieval results after fine-tuning are shown in Table \ref{tab:ft-msrvtt-didemo-lsmdc-msvd}. It can be observed that the conclusions drawn from zero-shot retrieval still hold true after fine-tuning.

\vspace{-0.3cm}
\begin{table}[]
\centering
\caption{Fine-tuned text-to-video retrieval results on MSRVTT, DiDeMo, LSMDC, and MSVD.}
\vspace{-0.2cm}
\resizebox{\textwidth}{!}{%
\begin{tabular}{ccccccccccccccccc}
\toprule
\multicolumn{1}{c|}{\multirow{2}{*}{Method}} & \multicolumn{4}{c|}{MSRVTT} & \multicolumn{4}{c|}{DiDeMo} & \multicolumn{4}{c|}{LSMDC} & \multicolumn{4}{c}{MSVD} \\
\multicolumn{1}{c|}{} & R@1$\uparrow$ &R@5 $\uparrow$ & R@10$\uparrow$ & \multicolumn{1}{c|}{MedR$\downarrow$} & R@1$\uparrow$ &R@5 $\uparrow$ & R@10$\uparrow$ & \multicolumn{1}{c|}{MedR$\downarrow$} & R@1$\uparrow$ & R@5$\uparrow$ & R@10$\uparrow$ & \multicolumn{1}{c|}{MedR$\downarrow$} & R@1$\uparrow$ & R@5$\uparrow$ & R@10$\uparrow$ & MedR$\downarrow$ \\ \midrule
\multicolumn{1}{c|}{Frozen \cite{bain2021frozen}} & 31.0 & 59.5 & 70.5 & \multicolumn{1}{c|}{3.0} & 31.0 & 59.8 & 72.4 & \multicolumn{1}{c|}{3.0} & 15.0 & 30.8 & 39.8 & \multicolumn{1}{c|}{20.0} & 45.6 & 79.8 & 88.2 & 2.0 \\
\multicolumn{1}{c|}{LaT \cite{bai2022lat}} & 35.3 & 61.3 & 72.9 & \multicolumn{1}{c|}{3.0} & 32.6 & 61.3 & 71.6 & \multicolumn{1}{c|}{3.0} & - & - & - & \multicolumn{1}{c|}{-} & 40.0 & 74.6 & 84.2 & 2.0 \\
\multicolumn{1}{c|}{RegionL \cite{yan2023video}} & 36.3 & 63.9 & 72.5 & \multicolumn{1}{c|}{3.0} & 32.5 & 60.8 & 72.3 & \multicolumn{1}{c|}{3.0} & 17.1 & 32.5 & 41.5 & \multicolumn{1}{c|}{18.0} & 44.0 & 74.9 & 84.3 & 2.0 \\
\multicolumn{1}{c|}{OA-Trans \cite{wang2022object}} & 35.8 & 63.4 & 76.5 & \multicolumn{1}{c|}{3.0} & 34.8 & 64.4 & 75.1 & \multicolumn{1}{c|}{3.0} & 18.2 & 34.3 & 43.7 & \multicolumn{1}{c|}{18.5} & 39.1 & 68.4 & 80.3 & 2.0 \\
\multicolumn{1}{c|}{MCQ \cite{ge2022bridging}} & 37.6 & 64.8 & 75.1 & \multicolumn{1}{c|}{3.0} & 37.0 & 62.2 & 73.9 & \multicolumn{1}{c|}{3.0} & 17.9 & 35.4 & 44.5 & \multicolumn{1}{c|}{15.0} & 52.0 & 82.8 & 90.0 & 1.0 \\
\multicolumn{1}{c|}{Miles \cite{ge2022miles}} & 37.7 & 63.6 & 73.8 & \multicolumn{1}{c|}{3.0} & 36.6 & 63.9 & 74.0 & \multicolumn{1}{c|}{3.0} & 17.8 & 35.6 & 44.1 & \multicolumn{1}{c|}{15.5} & \textbf{53.9} & 83.5 & 90.2 & 1.0 \\ 
\multicolumn{1}{c|}{TCP \cite{Zhang_2023_ICCV}} & 38.0 & 65.5 & 76.4 & \multicolumn{1}{c|}{7.0} & -- & -- & -- & \multicolumn{1}{c|}{--} & -- & -- & -- & \multicolumn{1}{c|}{--} & -- & -- & -- & -- \\ 
\rowcolor{mygray}
\multicolumn{1}{c|}{\textbf{EA-VTR}} & \textbf{39.5} & \textbf{67.2} & \textbf{77.0} & \multicolumn{1}{c|}{\textbf{2.0}} & \textbf{43.7} & \textbf{73.3} & \textbf{81.7} & \multicolumn{1}{c|}{\textbf{2.0}} & \textbf{22.0} & \textbf{40.8} & \textbf{51.0} & \multicolumn{1}{c|}{\textbf{10.0}} & 52.7 & \textbf{83.5} & \textbf{90.7} & \textbf{1.0} \\ \bottomrule
\end{tabular}%
}
\label{tab:ft-msrvtt-didemo-lsmdc-msvd}
\vspace{-0.5cm}
\end{table}

\vspace{-0.1cm}
\begin{table}[]
\centering
\scriptsize
\vspace{-0.5cm}
\caption{Zero-shot video action recognition results on UCF101 and HMDB51.}
\vspace{-0.2cm}
\resizebox{0.52\linewidth}{!}{%
\begin{tabular}{c|cccc|cccc}
\toprule
\multirow{2}{*}{Method} & \multicolumn{4}{c|}{UCF101} & \multicolumn{4}{c}{HMDB51} \\
 & S1 & S2 & S3 & Mean & S1 & S2 & S3 & Mean \\ \midrule
ClipBert \cite{lei2021less} & 27.5 & 27.0 & 28.8 & 27.8 & 20.0 & 22.0 & 22.3 & 21.4\\
Frozen \cite{bain2021frozen} & 45.4 & 44.7 & 47.7 & 45.9 & 27.5 & 28.3 & 27.7 & 27.8\\
MCQ \cite{ge2022bridging} & 51.1 & 54.3 & 53.8 & 53.1 & 38.0 & 36.1 & \textbf{39.1} & 37.7\\
Miles \cite{ge2022miles} & 51.8 & 53.4 & 52.8 & 52.7 & 38.4 & 38.6 & 37.8 & 38.3\\
\rowcolor{mygray}
\textbf{EA-VTR} & \textbf{60.0} & \textbf{62.4} & \textbf{60.5} & \textbf{61.0} & \textbf{38.6} & \textbf{39.8} & 37.9 & \textbf{38.8} \\ \bottomrule
\end{tabular}%
}
\label{tab:ucf-hmdb51}
\vspace{-0.3cm}
\end{table}

\noindent \textbf{Video Action Recognition.}
To verify the generalization ability of our method, we conduct zero-shot evaluation on video action recognition datasets UCF101 and HMDB51, by converting the video categories to captions with text prompts and uniformly sampling 16 frames for videos as input, as in previous works.
The results are presented in Table \ref{tab:ucf-hmdb51}, 
it can be seen that EA-VTR improves the average top-1 classification accuracy (Column ``Mean'' in the table) on UCF101 and HMDB51 by 8.3\% and 0.5\% compared to Miles \cite{ge2022miles}. 
This can be attributed to the fact that actions are inherently present in video events, and thus the event augmentation and learning method proposed in our work naturally enhances the understanding of actions.

\begin{table*}[htbp]
  \begin{minipage}[c]{0.6\textwidth}
    \centering
    \caption{Zero-shot Multi-event Video-Text Retrieval results on ActivityNet, with Recall@k-Average/One-Hit/All-Hit as metrics for each k = 1, 5, 10, 50 in Video-to-Text Retrieval and commonly used Recall@k as metric in Text-to-Video Retrieval. }
    \vspace{-0.1cm}
    \resizebox{\linewidth}{!}{%
    \begin{tabular}{>{\centering\arraybackslash}p{1.8cm}|cccc}
    \toprule
    \multirow{2}{*}{Method} & \multicolumn{4}{c}{Video-To-Text} \\
     & k=1 & k=5 & k=10 & k=50  \\ \midrule
      Frozen \cite{bain2021frozen} & 2.1/6.8/-- & 7.2/19.5/0.0 & 11.9/29.8/1.5 & 30.3/60.7/7.9 \\
      MCQ \cite{ge2022bridging}  & 2.8/8.8/-- & 9.7/25.2/1.0 & 15.0/36.8/2.2 & 35.9/68.9/10.4 \\
      Miles \cite{ge2022miles}  & 2.9/9.3/-- & 10.1/26.5/1.0 & 15.4/37.2/2.2 & 36.3/69.7/10.7 \\
      \rowcolor{mygray}
      \textbf{EA-VTR}  & $\pmb{4.4}$/$\pmb{13.9}$/-- & $\pmb{13.2}$/$\pmb{33.6}$/$\pmb{1.4}$ & $\pmb{19.8}$/$\pmb{45.8}$/$\pmb{3.5}$ & $\pmb{42.9}$/$\pmb{77.0}$/$\pmb{14.9}$ \\ \midrule
     & \multicolumn{4}{c}{Text-To-Video}  \\ 
     & k=1 & k=5 & k=10 & k=50 \\ \midrule
      Frozen \cite{bain2021frozen} & 6.9 & 19.4 & 28.4 & 53.7 \\
      MCQ \cite{ge2022bridging}  & 8.5 & 22.8 & 32.5 & 58.6 \\
      Miles \cite{ge2022miles}  & 8.6 & 22.9 & 32.4 & 58.1 \\
      \rowcolor{mygray}
      \textbf{EA-VTR} & \textbf{10.7} & \textbf{27.0} & \textbf{37.7} & \textbf{63.1} \\ \bottomrule
    \end{tabular}%
    }
    \label{tab:mevtr}
  \end{minipage}
  \vspace{-0.4cm}
  \hfill
  \begin{minipage}[c]{0.38\textwidth}
    \centering
    \caption{Fine-tuned Video Moment Retrieval results on ActivityNet. }
    \vspace{-0.2cm}
    \centering
    \footnotesize
    \resizebox{0.85\textwidth}{!}{%
    \begin{tabular}{@{}c|ccccc@{}}
    \toprule
    Method  & $\mathrm{R}@^{0.5}_1$ & $\mathrm{R}@^{0.7}_1$ & $\mathrm{R}@^{0.5}_5$ & $\mathrm{R}@^{0.7}_5$ \\ \midrule
    Frozen \cite{bain2021frozen} & 43.3 & 25.8 & 75.8 & 59.3 \\
    LocVTP \cite{cao2022locvtp} & 46.1 & 27.6 & 78.9 & 63.7 \\
    TCP \cite{Zhang_2023_ICCV} & 46.5 & 28.4 & 78.2 & 64.0 \\
    \rowcolor{mygray}
    \textbf{EA-VTR} & $\pmb{48.0}$ & $\pmb{29.6}$ & $\pmb{80.1}$ & $\pmb{65.0}$  \\ \bottomrule
    \end{tabular}%
    }
    \label{tab:moment-ret}
    
    \vfill 
    
    \scriptsize
    \vspace{0.35cm}
    \caption{Zero-shot Test of Time results on ActivityNet. 
             }
    \vspace{-0.2cm}
    \resizebox{0.8\linewidth}{!}{%
    \begin{tabular}{>{\centering\arraybackslash}p{1.5cm}|ccc}
    \toprule
    \multicolumn{1}{c|}{\multirow{2}{*}{Method}} & \multicolumn{3}{c}{ActivityNet} \\
     & Before & After & All \\ \midrule
    Frozen \cite{bain2021frozen} & 48.9 & 49.6 & 49.3  \\ 
    MCQ \cite{ge2022bridging} & 50.5 & 50.1 & 50.3 \\
    Miles \cite{ge2022miles} & 50.2 & 49.8 & 50.0 \\
    \rowcolor{mygray}
    \textbf{EA-VTR} & \textbf{62.6} & \textbf{60.5} & \textbf{61.6}   \\
     \bottomrule
    \end{tabular}%
    }
    \label{tab:sys_tl}
  \end{minipage}
\end{table*}

\noindent \textbf{Multi-event Video-Text Retrieval.}
The goal of Multi-event Video-Text Retrieval task is to recall all events related to a given video, which effectively reflects the comprehensive perception ability of the model for all video events. 
We perform zero-shot evaluation without any post-training on the ActivityNet dataset of the task by uniformly sampling 16 video frames as input, and the results are presented in Table \ref{tab:mevtr}.
On the video-to-text retrieval setting, it is evident that our method significantly outperforms others in all three metrics: Recall@k-Average/One-Hit/All-Hit, and the results for the text-to-video retrieval setting exhibit a similar trend. 
This validates that the ECL proposed in the work indeed benefits the ability of the model to capture all events within the video.

\noindent \textbf{Video Moment Retrieval.}
The purpose of video moment retrieval is to find the corresponding clip in the video given a text query, which can reflect the ability of the model to perceive a specific event.
We adapt our method for the video moment retrieval task on the ActivityNet dataset following the strategy in LocVTP. 
The results in Table \ref{tab:moment-ret} show that although our method is not specifically designed for this task, it still outperforms the customized SOTA methods LocVTP and TCP, achieving improvements of 1.9\% and 1.5\% compared to the two methods on $\mathrm{R}@^{0.5}_1$ respectively.

\noindent \textbf{Test of Time.}
Each sample in the Test of Time task contains a video and two video captions with reversed event orders, requiring the model to select the correct caption for the video.
This task aims to test the temporal understanding ability of the model for video events, and we performed zero-shot evaluation on the ActivityNet dataset provided by the authors.
Results in Table \ref{tab:sys_tl} indicate that it is a rather challenging task, as the performance of previous dual-encoder models is close to random guessing (50.0\%). 
It can be seen that our method has a significant advantage in determining the temporal relationships of events, exhibiting a 61.6\% time-order consistency across all samples, leading the second-place method by 11.3\%.
This demonstrates that ETL can effectively enhance the ability of the model to understand event temporal logic.

\subsection{Ablation Studies}

In this section, we conduct ablations to verify the effectiveness of our design choices through evaluating models for zero-shot text-to-video retrieval on MSRVTT and DiDeMo.
Due to the limited resources, we only use 1 million video-text pairs randomly selected from WebVid-2M for pre-training with a batch size of 512.

\begin{table}[]
\centering
\footnotesize
\caption{Ablation studies on the effectiveness and compatibility of the components in our method, including VTA, ECL and ETL.}
\vspace{-0.2cm}
\resizebox{0.9\linewidth}{!}{%
\begin{tabular}{>{\centering\arraybackslash}p{1.2cm}>{\centering\arraybackslash}p{1.2cm}>{\centering\arraybackslash}p{1.2cm}>{\centering\arraybackslash}p{1.2cm}|>{\centering\arraybackslash}p{1.5cm}>{\centering\arraybackslash}p{1.5cm}>{\centering\arraybackslash}p{1.5cm}|>{\centering\arraybackslash}p{1.5cm}>{\centering\arraybackslash}p{1.5cm}>{\centering\arraybackslash}p{1.5cm}}
\toprule
& \multirow{2}{*}{VTA} & \multirow{2}{*}{ETL} & \multirow{2}{*}{ECL} & \multicolumn{3}{c|}{MSRVTT} & \multicolumn{3}{c}{DiDeMo} \\
& &  &  & R@1$\uparrow$ & R@5$\uparrow$ & R@10$\uparrow$ & R@1$\uparrow$ & R@5$\uparrow$ & R@10$\uparrow$ \\ \midrule
A & \checkmark &  &  & 19.2 & 39.2 & 49.7 & 20.3 & 45.2 & 55.2 \\
B & \checkmark & \checkmark &  & 22.0 & 42.9 & 53.3 & 25.3 & 50.1 & 59.9 \\
C & \checkmark &  & \checkmark & 21.5 & 42.1 & 52.7 & 25.3 & 51.2 & 61.3  \\
D & \checkmark & \checkmark & \checkmark & \textbf{23.6} & \textbf{44.4} & \textbf{54.6} & \textbf{27.3} & \textbf{54.2} & \textbf{63.0} \\ \bottomrule
\end{tabular}%
}
\label{tab:module}
\vspace{-0.5cm}
\end{table}

\noindent \textbf{Are ECL and ETL effective and compatible?} 
Yes. 
From the results in Table \ref{tab:module}, we can make the following observations: 
(1) Models B and C both outperform baseline model A trained only with VTA. 
Specifically, model B, incorporating ECL, demonstrates a greater improvement on MSRVTT, while model C, integrating ETL, exhibits a more significant gain on DiDeMo. 
This suggests that the benefits of ECL and ETL are more prominent in event content and temporal learning.
(2) Furthermore, model D, incorporates ECL and ETL, achieves further improvements over models B and C on both datasets, suggesting that ECL and ECL are mutually compatible and beneficial.

\begin{table}[]
\centering
\vspace{-0.3cm}
\caption{Ablation studies on the effectiveness of Alternating Iteration Training.}
\vspace{-0.2cm}
\resizebox{0.7\linewidth}{!}{%
\begin{tabular}{>{\centering\arraybackslash}p{2.15cm}|>{\centering\arraybackslash}p{1.5cm}>{\centering\arraybackslash}p{1.5cm}>{\centering\arraybackslash}p{1.5cm}|>{\centering\arraybackslash}p{1.5cm}>{\centering\arraybackslash}p{1.5cm}>{\centering\arraybackslash}p{1.5cm}}
\toprule
\multirow{2}{*}{Method} & \multicolumn{3}{c|}{MSRVTT} & \multicolumn{3}{c}{DiDeMo} \\
 & R@1$\uparrow$ & R@5$\uparrow$ & R@10$\uparrow$ & R@1$\uparrow$ & R@5$\uparrow$ & R@10$\uparrow$ \\ \midrule
Baseline & 19.2 & 39.2 & 49.7 & 20.3 & 45.2 & 55.2\\ \midrule
w/o AIT. & 20.9 & 41.6 & 51.5 & 23.3 & 48.6 & 58.4 \\
with AIT. & \textbf{23.6} & \textbf{44.4} & \textbf{54.6} & \textbf{27.3} & \textbf{54.2} & \textbf{63.0} \\ \bottomrule
\end{tabular}%
}
\label{tab:ait}
\vspace{-0.5cm}
\end{table}

\noindent \textbf{Is Alternating Iteration Training effective?}
Yes.
Removing the Alternating Iterative Training (AIT) involves mixing the web-crawled and ETA-synthesized video-text pairs in the same batch for training. 
As can be seen from the comparisons in Table \ref{tab:ait}, the improvement obtained by removing AIT is significantly lower than that of our model, compared to the baseline model.
As previously mentioned in section \ref{sec:train_infer}, web-crawled and synthesized video-text pairs have different information densities, making them easily distinguishable. 
The contrast between them may not yield much gains, resulting in an effect similar to two independent contrastive learning with smaller batch sizes.

\vspace{-0.3cm}
\begin{table}[]
\centering
\caption{Event learning \vs textual augmentation. ``VC'' and ``FC'' are abbreviations for ``Video Caption`` and ``Frame Caption''.}
\vspace{-0.2cm}
\resizebox{\linewidth}{!}{%
\begin{tabular}{>{\centering\arraybackslash}p{1.2cm}>{\centering\arraybackslash}p{6.5cm}|>{\centering\arraybackslash}p{1.3cm}>{\centering\arraybackslash}p{1.3cm}>{\centering\arraybackslash}p{1.3cm}|>{\centering\arraybackslash}p{1.3cm}>{\centering\arraybackslash}p{1.3cm}>{\centering\arraybackslash}p{1.3cm}}
\toprule
& \multirow{2}{*}{Method} & \multicolumn{3}{c|}{MSRVTT} & \multicolumn{3}{c}{DiDeMo} \\
& & R@1$\uparrow$ & R@5$\uparrow$ & R@10$\uparrow$ & R@1$\uparrow$ & R@5$\uparrow$ & R@10$\uparrow$ \\ \midrule
A & VTA + VC & 19.2 & 39.2 & 49.7 & 20.3 & 45.2 & 55.2\\ 
B & VTA + Sample(VC, Concat(all FC)) & 21.1 & 41.8 & 51.9 & 23.8 & 48.8 & 58.9 \\ 
C & VTA + Concat(VC, FC) & 20.9 & 41.5 & 51.6 & 23.1 & 47.6 & 57.6 \\ \midrule
D & \textbf{EA-VTR} & \textbf{23.6} & \textbf{44.4} & \textbf{54.6} & \textbf{27.3} & \textbf{54.2} & \textbf{63.0} \\ \bottomrule
\end{tabular}%
}
\label{tab:other}
\vspace{-0.5cm}
\end{table}

\noindent \textbf{Does the benefit come from the proposed event learning  rather than merely textual augmentation?}
Yes. 
We reorganize the original web-crawled video captions and generated frame captions to train two text-augmented baselines with only video-level contrastive learning, \ie, VTA, including models B (sampling from video captions and concatenated frame captions), and C (concatenating video captions with frame captions).
As shown in Table \ref{tab:other}, text-augmented baselines exhibit better performance than the simple baseline A trained only with web-crawled video captions, but are still notably inferior to our EA-VTR with event learning, especially on DiDeMo with complex events.

\begin{table}[]
\centering
\vspace{-0.3cm}
\caption{Ablation studies on the event augmentation strategy.}
\vspace{-0.2cm}
\resizebox{0.9\linewidth}{!}{%
\begin{tabular}{>{\centering\arraybackslash}p{2.2cm}>{\centering\arraybackslash}p{2.2cm}|>{\centering\arraybackslash}p{1.5cm}>{\centering\arraybackslash}p{1.5cm}>{\centering\arraybackslash}p{1.5cm}|>{\centering\arraybackslash}p{1.5cm}>{\centering\arraybackslash}p{1.5cm}>{\centering\arraybackslash}p{1.5cm}}
\toprule
\multirow{2}{*}{Captioner} & \multirow{2}{*}{Decoding} & \multicolumn{3}{c|}{MSRVTT} & \multicolumn{3}{c}{DiDeMo} \\
 &  & R@1$\uparrow$ & R@5$\uparrow$ & R@10$\uparrow$ & R@1$\uparrow$ & R@5$\uparrow$ & R@10$\uparrow$ \\ \midrule
BLIP & BeamSearch & 21.7 & 42.1 & 52.3 & 23.9 & 51.7 & 60.2 \\
BLIP & Top-p & 23.6 & \textbf{44.4} & 54.6 & 27.3 & \textbf{54.2} & 63.0 \\
BLIP2 & Top-p & \textbf{23.9} & 44.0 & \textbf{54.7} & \textbf{27.8} & 53.2 & \textbf{63.3}  \\ \bottomrule
\end{tabular}%
}
\label{tab:captioner}
\vspace{-0.3cm}
\end{table}

\noindent \textbf{Can more diverse text improve downstream task generalization?}
Yes.
The diversity of frame captions can be manipulated by altering the decoding strategy of the captioner. 
Specifically, Table \ref{tab:captioner} shows that the model trained with less diverse frame captions obtained using deterministic BeamSearch \cite{freitag2017beam} decoding underperforms the model trained with non-deterministic Top-p decoded captions.
Non-deterministic decoding strategy provides different vocabulary or phrases to describe similar events, which aids in model generalization.

\noindent \textbf{Is a stronger image captioner like MLLM needed for event augmentation?} 
The recent exciting emergence of MLLM (Multi-Modal Large Language Model) \cite{li2023blip, zhu2023minigpt} raises this question, we thus answer it by replacing the captioner with the most commonly used MLLM, BLIP2. 
The results in Table \ref{tab:captioner} show that there is no significant improvement after the replacement, indicating that MLLM does not have a clear advantage in our method but requires longer computation time and memory. 
It also demonstrates that although our method relies on a captioner, it possesses good robustness to the ability of the captioner.

\begin{table}[]
\centering
\vspace{-0.3cm}
\caption{Ablations of video-text pair synthesis strategy in ETA. ``VWSum'', ``VConcat'', ``VCC'' and ``FCC'' are abbreviations for ``Video Weighted Summation'', ``Video Concatenation'', ``Video Caption concatenation'' and ``Frame Caption concatenation''.}
\vspace{-0.2cm}
\resizebox{\linewidth}{!}{%
\begin{tabular}{>{\centering\arraybackslash}p{1.2cm}>{\centering\arraybackslash}p{2.0cm}>{\centering\arraybackslash}p{2.0cm}|>{\centering\arraybackslash}p{1.5cm}>{\centering\arraybackslash}p{1.5cm}>{\centering\arraybackslash}p{1.5cm}|>{\centering\arraybackslash}p{1.5cm}>{\centering\arraybackslash}p{1.5cm}>{\centering\arraybackslash}p{1.5cm}}
\toprule
& \multirow{2}{*}{\makecell{Video\\Fusion}} & \multirow{2}{*}{\makecell{Text\\Fusion}} & \multicolumn{3}{c|}{MSRVTT} & \multicolumn{3}{c}{DiDeMo} \\
&  &  & R@1$\uparrow$ & R@5$\uparrow$ & R@10$\uparrow$ & R@1$\uparrow$ & R@5$\uparrow$ & R@10$\uparrow$ \\ \midrule
A & -- & -- & 19.2 & 39.2 & 49.7 & 20.3 & 45.2 & 55.2 \\ \midrule
B & VWSum & VCC & 19.5 & 39.7 & 49.9 & 20.6 & 44.9 & 55.7 \\
C & VConcat & VCC & 20.1 & 40.7 & 50.0 & 20.8 & 45.7 & 55.9 \\
D & VWSum & FCC & 21.3 & 41.8 & 52.3 & 23.3 & 50.3 & 60.3 \\
E & VConcat & FCC & \textbf{21.5} & \textbf{42.1} & \textbf{52.7} & \textbf{25.3} & \textbf{51.2} & \textbf{61.3} \\ \bottomrule
\end{tabular}%
}
\label{tab:etl}
\vspace{-0.3cm}
\end{table}

\noindent \textbf{Will the video-text synthesis strategy in ETA affect the gains of ETL?} 
Yes.
We compare different synthesis strategies in Table \ref{tab:etl}, and all models are trained only with VTA and ETL to form a transparent evaluation.
Model E in Table \ref{tab:etl} achieves a greater improvement compared to the baseline model A than model D, indicating that video concatenation is more suitable than weighted summation for synthesizing videos. 
This may be because the frames fused through weighted summation inevitably contain events from both videos (albeit in different proportions), leading to ambiguity in discriminating event temporal logic.
Additionally, models B and C, which concatenate web-crawled video captions for the synthesized videos, do not exhibit as significant gains as models D and E. 
This validates that the web-crawled captions contain less event information, making them less conducive to event learning.

\subsection{Qualitative Analysis}

\begin{figure*}
    \centering
    \vspace{-0.3cm}
    \includegraphics[width=\textwidth]{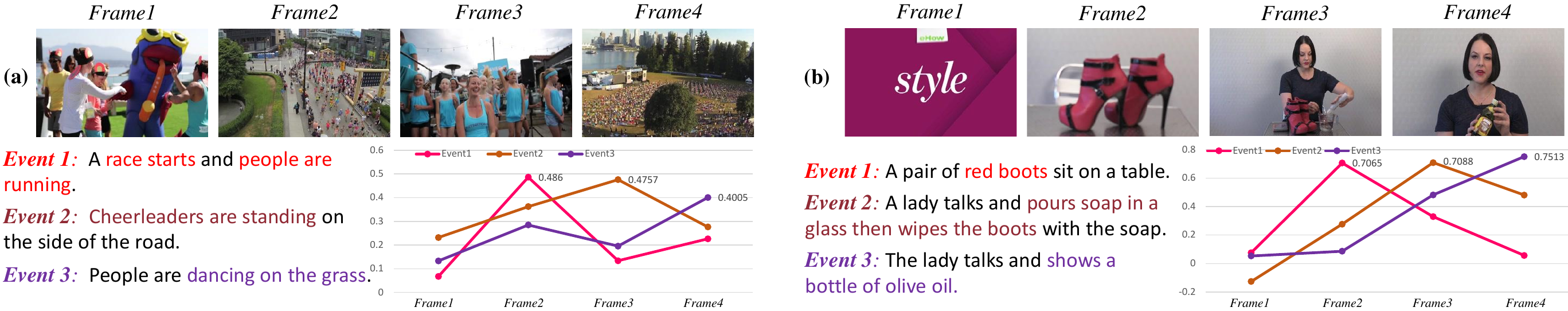}
    \vspace{-0.3cm}
    \caption{
    Examples of frame-event alignment:
    Above are the extracted video frames, while the bottom left shows multiple events occurring at different times in the video, distinguished by different colors (with key event information in the text also colored). 
    The bottom right displays the similarity score curves between the text features of these events and the visual features of video frames.
    }
    \label{fig:frame_sim}
    \vspace{-0.3cm}
\end{figure*}

\noindent \textbf{Frame-Event Alignment.}
To demonstrate that the frame-level visual representations optimized by ECL are meaningful, we visualize the frame-event alignment relationships in Figure \ref{fig:frame_sim}, with samples sourced from ActivityNet. 
Observing Figure \ref{fig:frame_sim} (a), we can find that the events and video frames exhibit the correct corresponding relationships, such as event 1 ``A race starts and people are running'' with frame 2.
Additionally, some challenging frame-event alignment relationships can also be accurately captured by the model. 
As shown in Figure \ref{fig:frame_sim} (b), where the event subjects of events 2 and 3 are the same woman, with the difference being the items she is presenting.
Nonetheless, the model correctly associates event 2 with frame 3 and event 3 with frame 4.

\section{Conclusion}
In this work, we are dedicated to enhancing the ability of video-text retrieval model to comprehensively understand the video events, thereby enabling a event-aware retriever. 
To achieve this goal, we have made efforts to supplement missing event information in the pre-training data and improve the model's ability to capture detailed and complex video-text event alignment.
On this basis, we validate the proposed method by thorough experimentation from multiple perspectives.
In the future, we will continue to explore the help of event learning for other video-text tasks such as video question answering and video captioning. \textbf{Limitations: } Due to resource constraints, we are unable to further expand the model and data scale to compare with models like InternVideo \cite{wang2022internvideo}.

\clearpage

\bibliographystyle{splncs04}
\bibliography{main}

\clearpage

\appendix
\renewcommand{\thesection}{\Alph{section}}
\section{Details of Downstream Datasets and Evaluation Metrics.}

\subsection{Text-To-Video Retrieval}
The text-to-video retrieval task consists of four downstream datasets: \textbf{MSRVTT} \cite{xu2016msr}, \textbf{DiDeMO} \cite{anne2017localizing}, \textbf{LSMDC} \cite{rohrbach2015dataset} and \textbf{MSVD} \cite{chen2011collecting}.
\textbf{MSRVTT} consists of 10K YouTube videos, including 9K videos for training and 1K videos for testing, each video is equipped with 20 descriptions.
\textbf{DiDeMo} contains 10K Flickr videos with 40K descriptions, where the test set also has 1K videos. 
As in previous work \cite{bain2021frozen, ge2022bridging, ge2022miles}, we concatenate all descriptions of a video into a single text query for evaluating paragraph-to-video retrieval.
The videos for \textbf{LSMDC} are 118,081 clips from 202 movies, where the test set still contains 1K video clips from movies that do not overlap with the training and validation sets.
\textbf{MSVD} contains 1,970 YouTube videos with 80K descriptions, and we adopt the standard split \cite{liu2019use, patrick2020support} of 1200, 100, and 670 videos for training, validation, and testing.

We report the widely-used R@k (k=1,5,10) for text-to-video retrieval, which measures the proportion of samples for which the correct match for the text query can be found in the top-k retrieved results.
In addition, we also report the MedR for text-to-video retrieval, which is the median rank of the correct match for the text query in the sorted results.

\subsection{Video Action Recognition}
The video action recognition task consists of two datasets: \textbf{UCF101} \cite{soomro2012ucf101} and \textbf{HMDB51} \cite{kuehne2011hmdb}.
\textbf{UCF101} contains 13,320 videos in 101 action categories, and \textbf{HMDB51} contains 6,766 videos in 51 action categories. 
The test sets of both datasets are divided into three parts: S1, S2 and S3.

We report top-1 classification accuracy for the video action recognition task, defined as the ratio of the number of correctly classified videos to the total number of the test videos.

\subsection{Multi-event Video-Text Retrieval}
The Multi-event Video-Text Retrieval task is evaluated on the \textbf{ActivityNet} \cite{caba2015activitynet} dataset.
\textbf{ActivityNet} contains video-text pairs crawled from YouTube, which covers a wide range of daily scenarios, with a training set containing 10,009 videos and a publicly accessible test set "val1" containing 4,917 videos/17,505 captions. 
Each video has multiple event captions annotated with time intervals, allowing us to construct a test scenario for Multi-event Video-Text Retrieval by treating all captions of the video as its related events.

Recall@k-Average/One-Hit/All-Hit \cite{zhang2023multi} are employed for the Multi-event Video-Text Retrieval task.
(1) Recall@k-Average refers to the proportion of the correct matches in the top-k ranked retrieved events.
(2) Recall@k-One-Hit represents whether any of the correct matches are in the top-k ranked retrieved events.
(3) Recall@k-All-Hit indicates whether all correct matches can be included within the top-k ranked retrieved events.
All these metrics are averaged across the entire test set.

\vspace{-0.3cm}
\subsection{Video Moment Retrieval}
The video moment retrieval task utilizes the captions and corresponding time interval annotations provided in \textbf{ActivityNet} as training and test data, with the goal of finding the corresponding video clip for a given caption query.

Following previous works \cite{cao2022locvtp, Zhang_2023_ICCV}, we use $\mathrm{R}@^{\theta}_n$ as the evaluation metric for the task, which represents the percentage of cases where at least one of the top-$n$ retrieved clips has an IOU greater than $\theta$ with the ground-truth clip. 
In our work, we set $n \in \{1, 5\}$ and $\theta \in \{0.5, 0.7\}$.

\subsection{Test of Time}
The Test of Time task is also evaluated on the \textbf{ActivityNet} dataset.
For a video in \textbf{ActivityNet}, two selected event captions are concatenated into a single sentence by the conjunction "before/after" based on the temporal order of the events, while constructing a swapped sentence with the opposite event order.
The task goal is to identify the sentence that correctly matches the temporal order of events in the video.

We follow previous work \cite{bagad2023test} and use time-order consistency as the evaluation metric for the Test of Time task, \ie, the proportion of videos for which the model correctly associates text that is time order consistent with the video.

\section{Mining of Single-Event Videos}
To validate our observation that a considerable portion of WebVid \cite{bain2021frozen} videos contain only a single event, we have designed a quantitative calculation method to determine whether a video $V_i$ is a single-event video. 
Specifically, we first uniformly sample $K$ frames $\{v_1, v_2, ..., v_K\}$ from $V_i$ and then use a pre-trained image encoder, \ie, CLIP, to extract their respective frame features $\{f_1, f_2, ..., f_k\}$. 
We iterate through $k = 2, ..., K$ to compute the inter-frame cosine similarity $s_{1k} = \mathrm{cosine}(f_1, f_k)$ between the $k$-th frame and the first frame.
When a video contains multiple events, there will be event transitions, meaning that the cosine similarity between a certain frame and the first frame is lower than a predefined threshold $\tau$. We can determine whether the video contains only a single event by checking for the presence of such cases, defined as follows:
\begin{equation}
    V_i \in \left\{ 
    \begin{aligned}
        \mathrm{Single-event \ Video} & , \quad \mathrm{min}(\{s_{1k}|k=2,...,N\}\}) \geq \tau \\
        \mathrm{Multi-event \ Video} & , \quad \mathrm{min}(\{s_{1k}|k=2,...,N\}\}) < \tau
    \end{aligned}.
    \right.
\end{equation}

\section{Implementations}
Multi-Granularity Video Encoder contains 12 blocks with patch size P = 16 and hidden state dimension D = 768.
The dimensions of the embedding space for ECL, ETL and VTA are all set to 256.
The temperature hyper-parameters for ECL, ETL and VTA are set to 0.07, 0.07 and 0.05.
The above implementation details follow the recent works \cite{bain2021frozen, ge2022bridging, ge2022miles} for fair comparison. 

For the Test of Time task, we find that the text side of the current retrieval model does not seem to understand relational words, such as "before/after", very well. 
Therefore, we dismantle the relational words and concatenate the two event captions according to the temporal relationship during testing. 

\section{Ablation Studies}

\begin{table}[]
\centering
\vspace{-0.5cm}
\caption{Ablation studies on the number of frames used for ECA.}
\vspace{-0.2cm}
\footnotesize
\resizebox{0.5\linewidth}{!}{%
\begin{tabular}{@{}>{\centering\arraybackslash}p{2.15cm}|ccc|ccc@{}}
\toprule
\multirow{2}{*}{\makecell{Augmented \\ Frames}} & \multicolumn{3}{c|}{MSRVTT} & \multicolumn{3}{c}{DiDeMo} \\
&  R@1 & R@5 & R@10 & R@1 & R@5 & R@10 \\ \midrule 
-- & 19.2 & 39.2 & 49.7 & 20.3 & 45.2 & 55.2 \\ \midrule
1 Frame & 20.8 & 41.4 & 51.6 & 23.6 & 48.1 & 58.6  \\
2 Frames & 21.6 & 42.4 & 52.2 & 24.5 & 49.3 & 59.3 \\
3 Frames & 21.8 & \textbf{43.0} & 52.9 & 24.9 & 49.7 & 59.8 \\
4 Frames & \textbf{22.0} & 42.9 & \textbf{53.3} & \textbf{25.3} & \textbf{50.1} & \textbf{59.9} \\ \bottomrule
\end{tabular}%
}
\label{tab:ecl}
\vspace{-0.3cm}
\end{table}

\noindent \textbf{Will the amount of event content information augmented by ECA affect the gain of ECL?} 
Yes. 
We can control the amount of augmented event content information available in ECL by adjusting the number of generated frame-level image-text pairs in ECA.
From Table \ref{tab:ecl}, we can intuitively observe that as the number of image-text pairs increases, corresponding to the increase in event content information, the performance of the model also improves accordingly.
At the same time, when the number of image-text pairs generated for each video changes from 3 to 4, the performance exhibits only a slight variation. 
This suggests that using 4 frames for ECA is sufficient for short videos.

\begin{table}[]
\centering
\vspace{-0.3cm}
\caption{Ablation studies on the capacity of the text encoder.}
\vspace{-0.2cm}
\footnotesize
\resizebox{0.5\linewidth}{!}{%
\begin{tabular}{@{}>{\centering\arraybackslash}p{2.15cm}|ccc|ccc@{}}
\toprule
\multirow{2}{*}{\makecell{Text Encoder}} & \multicolumn{3}{c|}{MSRVTT} & \multicolumn{3}{c}{DiDeMo} \\
&  R@1 & R@5 & R@10 & R@1 & R@5 & R@10 \\ \midrule 
DistilBERT & 26.3 & 49.3 & 60.2 & 29.9 & 56.6 & 67.3  \\
BERT & \textbf{28.0} & \textbf{53.1} & \textbf{62.3} & \textbf{32.7} & \textbf{58.9} & \textbf{68.9} \\ \bottomrule
\end{tabular}%
}
\label{tab:text_encoder}
\vspace{-0.3cm}
\end{table}

\noindent \textbf{Does the capacity of the text encoder affect the performance?}
Yes.
Previous works \cite{bain2021frozen, ge2022bridging, ge2022miles} have employed DistilBERT \cite{sanh2019distilbert} as the text side of the retrieval model, as its capacity is sufficient to accommodate all text under the typical pre-training settings of CC3M combined with WebVid-2M. 
However, the introducing of more text in the event-augmented pre-training data has led us to find that the capacity of DistilBERT is no longer adequate, we thus replace it with the larger-capacity BERT \cite{devlin2018bert}.
The comparisons can be seen in Table \ref{tab:text_encoder}, where it is evident that using BERT as the text side of the retrieval model yields significantly better performance.

\section{Visualizations}

\subsection{Event Content Augmentation}

\begin{figure}
    \centering
    \includegraphics[width=0.65\linewidth]{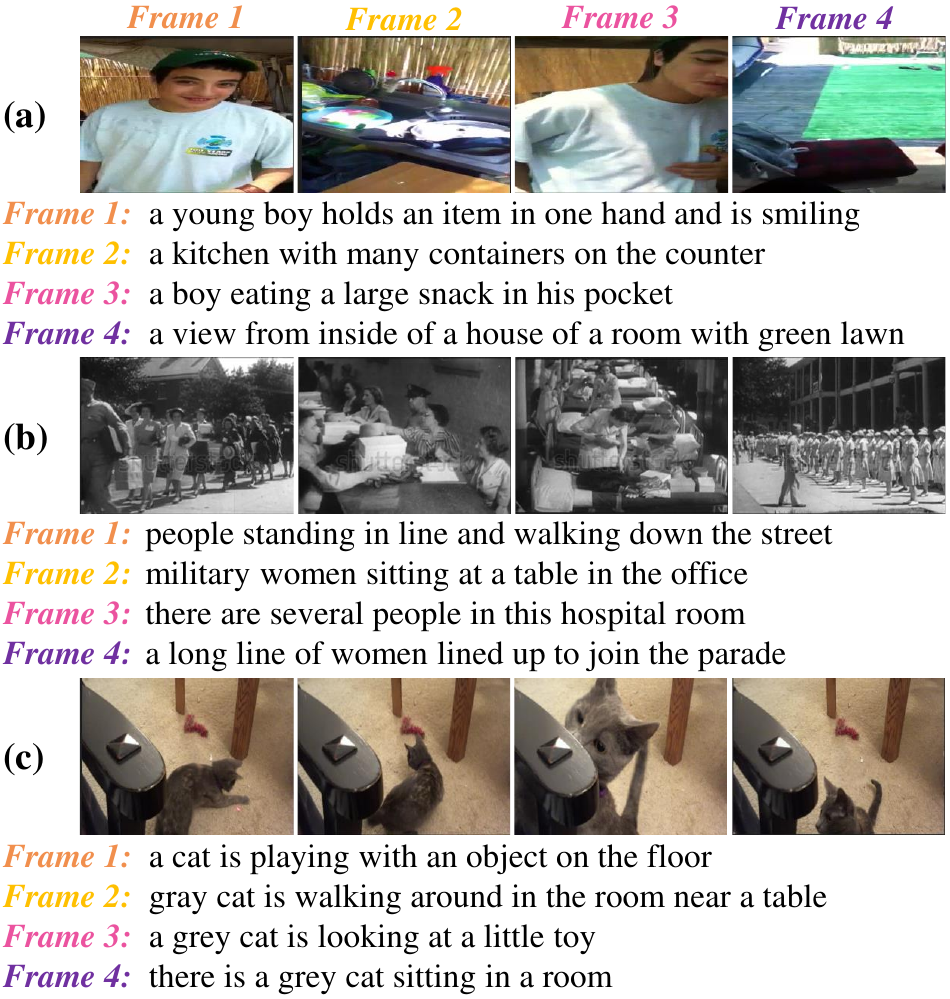}
    \caption{
    Examples of ECA. The top of each example are video frames extracted from 4 evenly divided clips of the video (consistent with the input of the video encoder), and below are event captions generated with the image captioner for each video frame.
    }
    \vspace{-0.2cm}
    \label{fig:ecl}
    \vspace{-0.3cm}
\end{figure}

We provide examples of ECA in Figure \ref{fig:ecl}. 
As observed, for videos like (a) and (b) with obvious event transitions, the existing image captioner can provide suitable event captions for each video frame, such as ``a young boy holds an item in one hand and is smiling" for Frame 1 of video (a) and ``military women sitting at a table in the office'' for Frame 2 of video (b). 
For videos like (c), where the event changes throughout the video are less noticeable, the non-deterministic generation strategy, \ie, Top-P sampling, can also offer relatively different event captions for visually similar but distinct video frames, especially the Frame 1 and Frame 2 of video (c), thereby increasing the diversity of event captions and improving the generalization ability of the model.

\subsection{Text-to-Video Retrieval}

\begin{figure}
    \centering
    \includegraphics[width=0.6\linewidth]{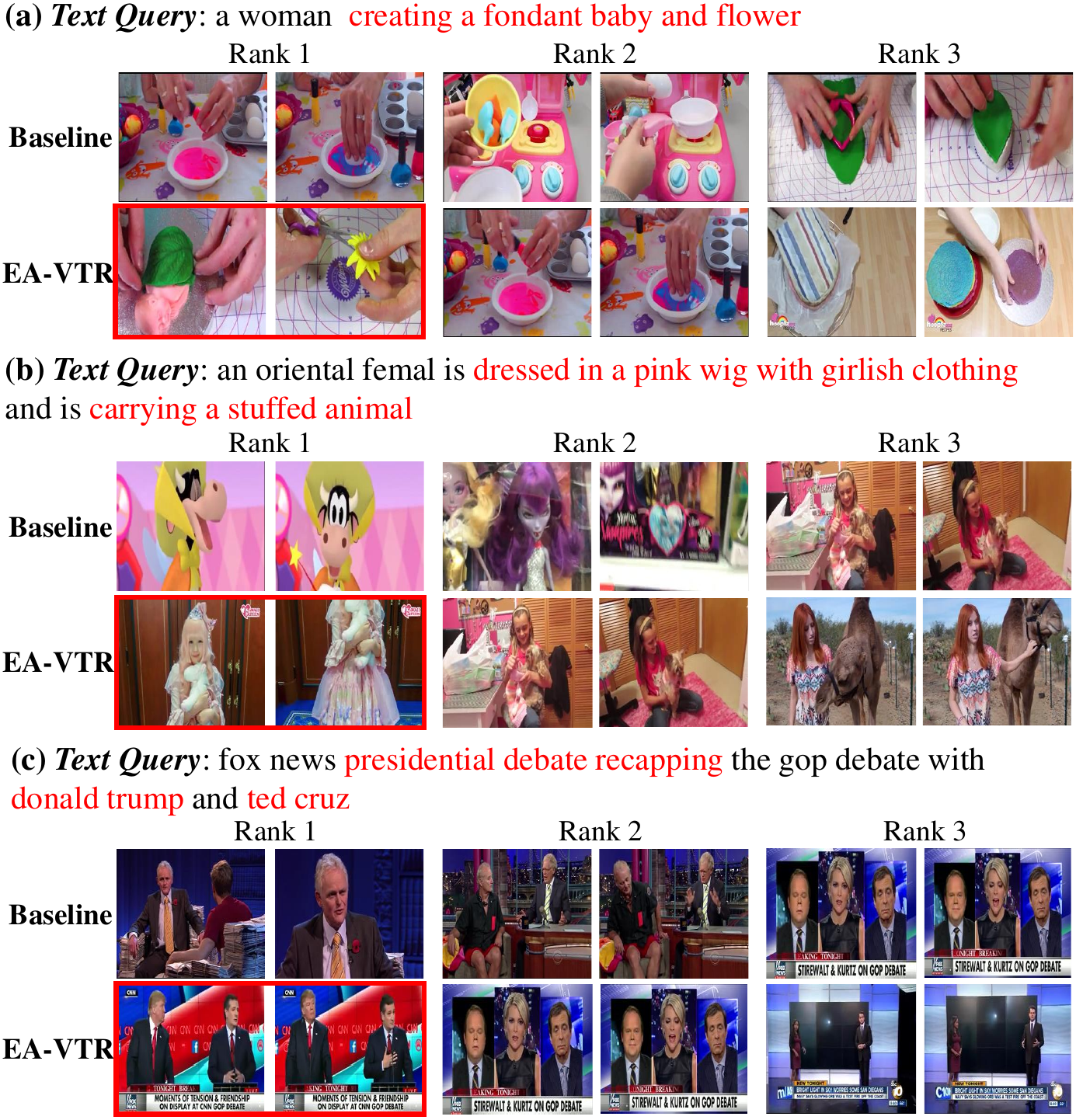}
    \caption{
    Illustrations of text-to-video retrieval results for the baseline model and our EA-VTR. 
    The top-3 ranked videos are provided for each text query, and 2 representative frames are chosen to represent each video, where the key parts of the text query are highlighted in red and the ground-truth video for each text query is in the red box.
    }
    \label{fig:t2v}
    \vspace{-0.3cm}
\end{figure}

We visualize the comparisons of text-to-video retrieval results between the baseline model and our EA-VTR in Figure \ref{fig:t2v}. 
Firstly, it can be observed that our method demonstrates a better understanding of common events. 
For instance, our method can simultaneously understand both "creating a fondant baby" and "creating a flower" events in the text query of (a) and retrieve the correct video, whereas the baseline model does not find any video that fully matches either event. 
Moreover, our method exhibits better ranking ability for partially relevant negative videos, such as in (b), where the video ranked second by our approach is relatively consistent with the events "dressed in a pink wig" and "carrying a stuffed animal." 
However, the baseline method ranks it third, behind a less relevant video.
Lastly, for events involving specific individuals, such as the debate between "Donald Trump" and "Ted Cruz" in (c), our method still achieves accurate understanding and retrieves the ground truth video.

\subsection{Multi-event Video-Text Retrieval}

\begin{figure*}
    \centering
    \includegraphics[width=\textwidth]{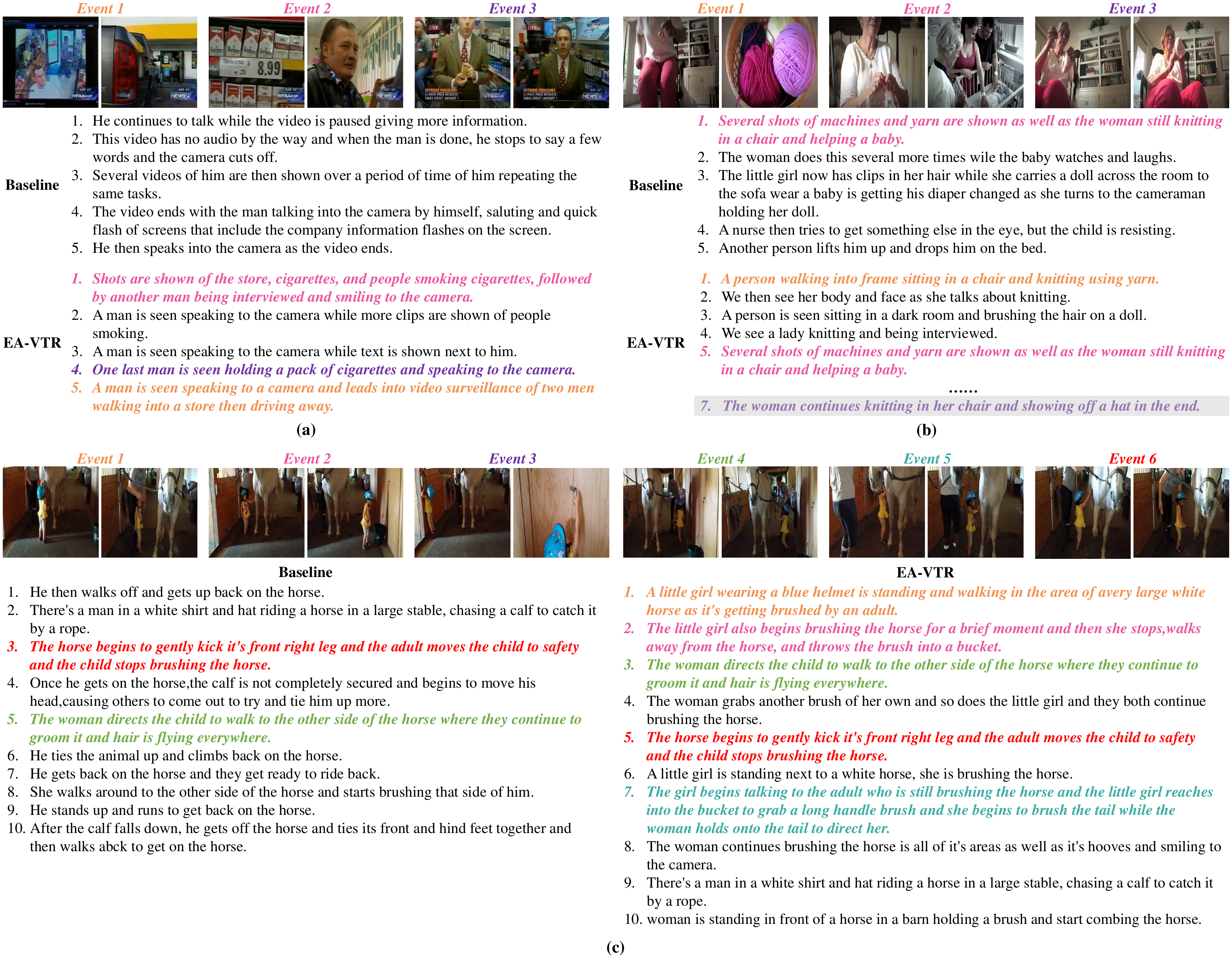}
    \caption{
    Illustrations of Multi-event Video-Text Retrieval results for the baseline model and our EA-VTR.
    The top of each example presents visual information about multiple events occurring in the video, with 2 representative frames are chosen to represent each event, and the bottom shows the top-K ranked event captions retrieved by the different models. 
    Different events are distinguished by different colors, and the same color is also marked in the text below if the corresponding event is captured correctly.
    }
    \label{fig:mevtr}
    \vspace{-0.3cm}
\end{figure*}

The visualizations of the qualitative comparisons for Multi-event Video-Text Retrieval results are shown in Figure \ref{fig:mevtr}. 
It is evident that our method achieves a more comprehensive understanding of video events, as in (a), where all three ground-truth events are included in the top-5 ranked captions of our approach, while the top-5 ranked captions of the baseline model do not contain any ground-truth events.
Even for videos like (b), where multiple events are quite similar, our method includes 2/3 ground truth events in the top-5 ranked captions, and the third event is also captured in the 7-th ranked caption. 
In contrast, the top-5 ranked captions of the baseline model only contain 1/3 ground truth events.
As the videos become more complex, \ie, containing more events, such as in (c), the advantages of our method become more apparent. 
Our approach includes 5/6 ground-truth events in the top-10 ranked captions, while the results of baseline model only contain 2/6 ground-truth events.

\subsection{Test of Time}

\begin{figure}
    \centering
    \includegraphics[width=0.6\linewidth]{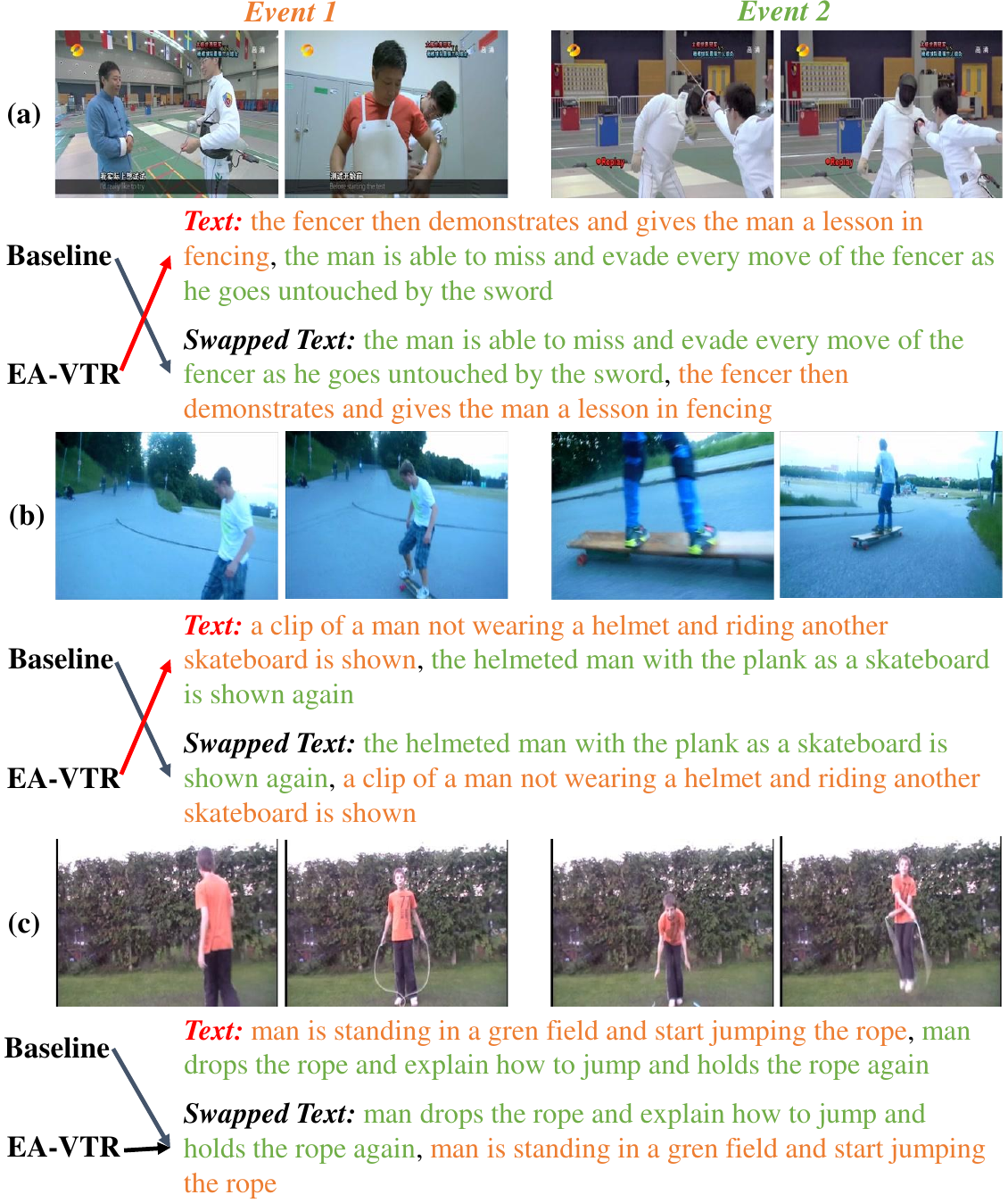}
    \caption{
    Illustrations of Test of Time results for the baseline model and our EA-VTR.
    The top of each example presents two video events that occur in a chronological order, with 2 representative frames are chosen to represent each event, and the bottom presents the text of the two events occurring in the correct order as well as the swapped text in the opposite event order, and the text selected by the different models to match the video.
    }
    \label{fig:tot}
    \vspace{-0.3cm}
\end{figure}

In Figure \ref{fig:tot}, we provide visualization examples of qualitative comparisons between the baseline model and our EA-VTR on the Test of Time task. 
It can be seen that the baseline model cannot understand the temporal relationship between events, even for cases like (a) with apparent transitions. 
In contrast, our method can not only correctly determine the temporal relationship between events in (a) but also accurately identify the temporal relationship between two visually similar events in (b). 
However, for cases like (c), where even humans find it difficult to discern the differences between two events based solely on visual information, our method is unable to succeed.

\vspace{-0.3cm}

\end{document}